\pdfoutput=1
\documentclass[11pt]{article}

\usepackage[final]{acl}

\usepackage{times}
\usepackage{latexsym}

\usepackage[T1]{fontenc}
\usepackage[utf8]{inputenc}

\usepackage{microtype}

\usepackage{inconsolata}

\usepackage{graphicx}

\usepackage{booktabs}
\usepackage{url}

\usepackage{tcolorbox}
\usepackage{xcolor}

\usepackage{multirow}

\usepackage{amsmath}
\usepackage{amssymb}

\tcbset{
  aibox/.style={
    width=\linewidth,
    top=7pt,
    bottom=2pt,
    colback=blue!6!white,
    colframe=black,
    colbacktitle=black,
    enhanced,
    center,
    attach boxed title to top left={yshift=-0.1in,xshift=0.15in},
    boxed title style={boxrule=0pt,colframe=white,},
  }
}
\newtcolorbox{AIbox}[2][]{aibox,title=#2,#1}

\usepackage{enumitem}

\setlist[itemize]{topsep=3pt, partopsep=0pt, parsep=0pt, itemsep=3pt}

\title{Through the Valley: Path to Effective Long CoT Training for\\Small Language Models}

\author{First Author \\
  Affiliation / Address line 1 \\
  Affiliation / Address line 2 \\
  Affiliation / Address line 3 \\
  \texttt{email@domain} \\\And
  Second Author \\
  Affiliation / Address line 1 \\
  Affiliation / Address line 2 \\
  Affiliation / Address line 3 \\
  \texttt{email@domain} \\}

\author{
 \textbf{Renjie Luo}\textsuperscript{1} \;\;\;  
 \textbf{Jiaxi Li}\textsuperscript{1} \;\;\;  
 \textbf{Chen Huang}\textsuperscript{1} \;\;\;  
 \textbf{Wei Lu}\textsuperscript{2}
\\
\textsuperscript{1}Singapore University of Technology and Design, \textsuperscript{2}Nanyang Technological University
 \\
 \texttt{renjie.luo@outlook.com}, \texttt{wei.lu@ntu.edu.sg} \\
}

\begin{document}
\maketitle

\begin{abstract}
  Long chain-of-thought (CoT) supervision has become a common strategy to enhance reasoning in language models. While effective for large models, we identify a phenomenon we call \emph{Long CoT Degradation}, in which small language models (SLMs; $\leq$3B parameters) trained on limited long CoT data experience significant performance deterioration. Through extensive experiments on the Qwen2.5, LLaMA3 and Gemma3 families, we demonstrate that this degradation is widespread across SLMs. In some settings, models trained on only 8k long CoT examples lose up to 75\% of their original performance before fine-tuning. Strikingly, we further observe that for some particularly small models, even training on 220k long CoT examples fails to recover or surpass their original performance prior to fine-tuning. Our analysis attributes this effect to error accumulation: while longer responses increase the capacity for multi-step reasoning, they also amplify the risk of compounding mistakes. Furthermore, we find that \emph{Long CoT Degradation} may negatively impacts downstream reinforcement learning (RL), although this can be alleviated by sufficiently scaled supervised fine-tuning (SFT). Our findings challenge common assumptions about the benefits of long CoT training for SLMs and offer practical guidance for building more effective small-scale reasoning models.
\end{abstract}

\begin{figure}[ht]
    \centering
    \includegraphics[width=0.9\linewidth]{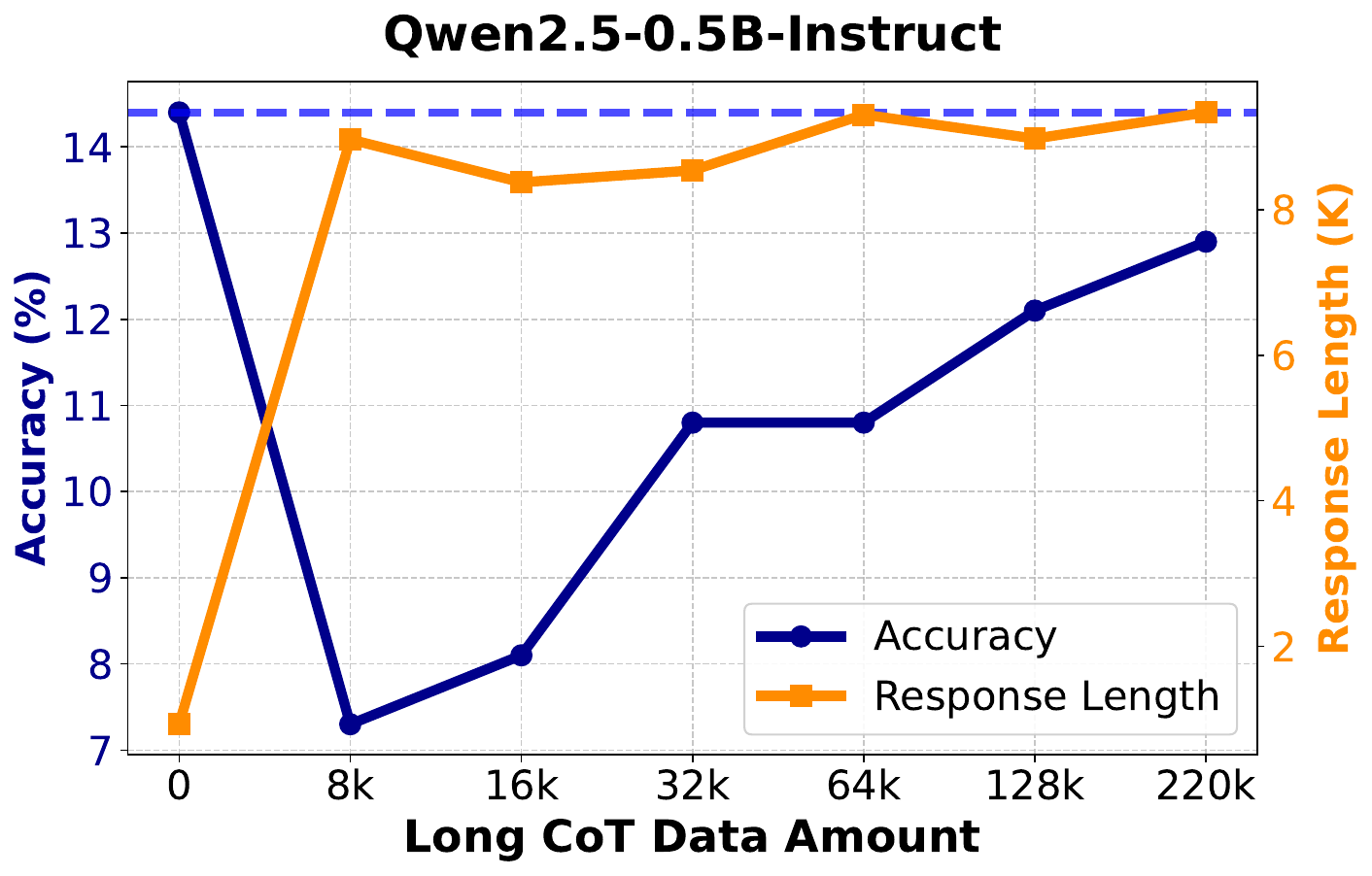}
    \vspace{-0.5em}
    \caption{Accuracy and response length for Qwen2.5-0.5B across varying amounts of long CoT SFT data. Performance drops markedly at smaller data scales (8k-16k), even as response length increases significantly, indicating a critical failure mode in which the model generates longer but less accurate reasoning traces. We term this phenomenon \emph{Long CoT Degradation}.}
    \vspace{-0.5em}
    \label{fig:degradation_05B}
\end{figure}

% 下面是被删掉的第二段
% Among these LRMs, DeepSeek-R1~\cite{deepseek-r1} stands out as the only fully open-source model within the top-performance tier. Consequently, it has become a popular choice as a teacher model for supervised fine-tuning (SFT) datasets~\cite{openr1,zhao20251,bespoke_stratos,slam-distillation-from-r1,openthoughts}. Models distilled from DeepSeek-R1 data, and further enhanced by reinforcement learning (RL), consistently achieve state-of-the-art (SOTA) performance within their respective size categories~\cite{deepscaler2025,deepcoder2025,nvidiaAcemath2024}.

\section{Introduction}

Large reasoning models, such as OpenAI-o3~\cite{openai2025o3o4}, Kimi-k1.5~\cite{kimik15}, and DeepSeek-R1~\cite{deepseek-r1} have recently demonstrated impressive capabilities in complex reasoning tasks. A key strength of these models is their ability to generate long chain-of-thought (CoT) responses, which usually demonstrate advanced, reflective reasoning behaviors.
These detailed reasoning responses, referred to as long CoT, constitute valuable resources for enhancing the reasoning ability of large language models (LLMs).
% These detailed reasoning responses, referred to as long CoT data, serve as valuable resources for training and enhancing reasoning capabilities.

% Within the domain of long CoT generation, existing research has primarily explored two settings: the use of small but carefully curated long CoT datasets (typically around 1k examples) to train large-scale models (e.g., 32B parameters)~\cite{muennighoff2025s1,ye2025limo}; and studies that conduct supervised fine-tuning using moderately sized datasets (approximately 8k examples) to train smaller models~\cite{li2025small,yeo2025demystifying}, often relying on LRM-generated CoTs. These practices, while effective in certain contexts, leave open questions about the broader reliability and scalability of long CoT supervision, particularly for small language models (SLMs; e.g., models with $\leq$8B parameters).

% Currently, long CoT data is commonly used in two ways: (1) in limited amounts ($\leq$10k examples) for cold-start supervised fine-tuning (SFT) as a warm-up to reinforcement learning (RL)~\cite{deepseek-r1,openr1,chen2025empirical}, and (2) at scale ($>$100k examples) for knowledge distillation~\cite{hinton2015distilling}, to enhance the reasoning abilities of general-purpose LLMs~\cite{deepseek-r1,openr1,wen2025light,bercovich2025llamanemotronefficientreasoningmodels}. Despite their widespread use, the choice of data scale in both settings is largely heuristic, and there is little empirical analysis of how long CoT data quantity affects model performance across training paradigms. 
Despite the growing use of long CoT data in LLM training, there remains no consistent strategy for its integration. Current practices vary widely across models and objectives. For instance, some studies show that even supervised fine-tuning (SFT) with a relatively small amount of long CoT data (e.g., $\leq$10k examples) can effectively enhance reasoning capabilities in LLMs~\citep{muennighoff2025s1, xu2025redstar, ye2025limo}. Others combine either limited or extensive long CoT SFT with subsequent RL training~\citep{openr1, chen2025empirical, bercovich2025llamanemotronefficientreasoningmodels, wen2025light, deepseek-r1}. However, in almost all these cases, the choice of data scale tends to be heuristic, and currently, there is limited empirical understanding of how the scale of long CoT data influences model performance across different training paradigms and different model sizes. 

The call for a closer examination of this underexplored topic is especially pertinent given the growing interest in developing and deploying small language models (SLMs), where strong reasoning capabilities are crucial due to their rising popularity and prevalence~\citep{li2023symbolic, hui2024smaller, lee2024mentor, agarwal2024policy}. Compared to larger models, SLMs typically have limited capacity, which may affect their ability to generalize from verbose supervision, such as long CoT data~\citep{feng2024teaching}. Although effective for large models, long CoT's verbosity may overwhelm smaller models, making it unclear to what extent SLMs can benefit from such training. Recent work even suggests that short CoT may be more effective for SLMs~\citep{li2025small}. However, these findings are based on relatively small-scale datasets (e.g., 8k examples), leaving it uncertain whether they hold at larger data scales.

%This gap in standard practice is especially pertinent for small language models (SLMs), where the effects of long CoT supervision might differ significantly from those observed in larger models -- a topic that remains largely underexplored. 

% This gap is especially critical for small language models (SLMs), which are increasingly valued for their efficiency and deployability.

% This gap is particularly relevant for small language models (SLMs), which are increasingly valued for efficient deployment. While long CoT data has been primarily studied in large models with greater reasoning capacity, the high inference cost of lengthy outputs makes it important to understand the effects of long CoT on SLMs. Notably, recent work suggests that short CoT may even outperform long CoT for SLMs~\cite{li2025small}; however, these findings are based on a limited dataset (8k examples), and it remains unclear whether this trend holds as the scale of long CoT data increases.

\begin{figure*}[ht]
    \centering
    \includegraphics[width=1\linewidth]{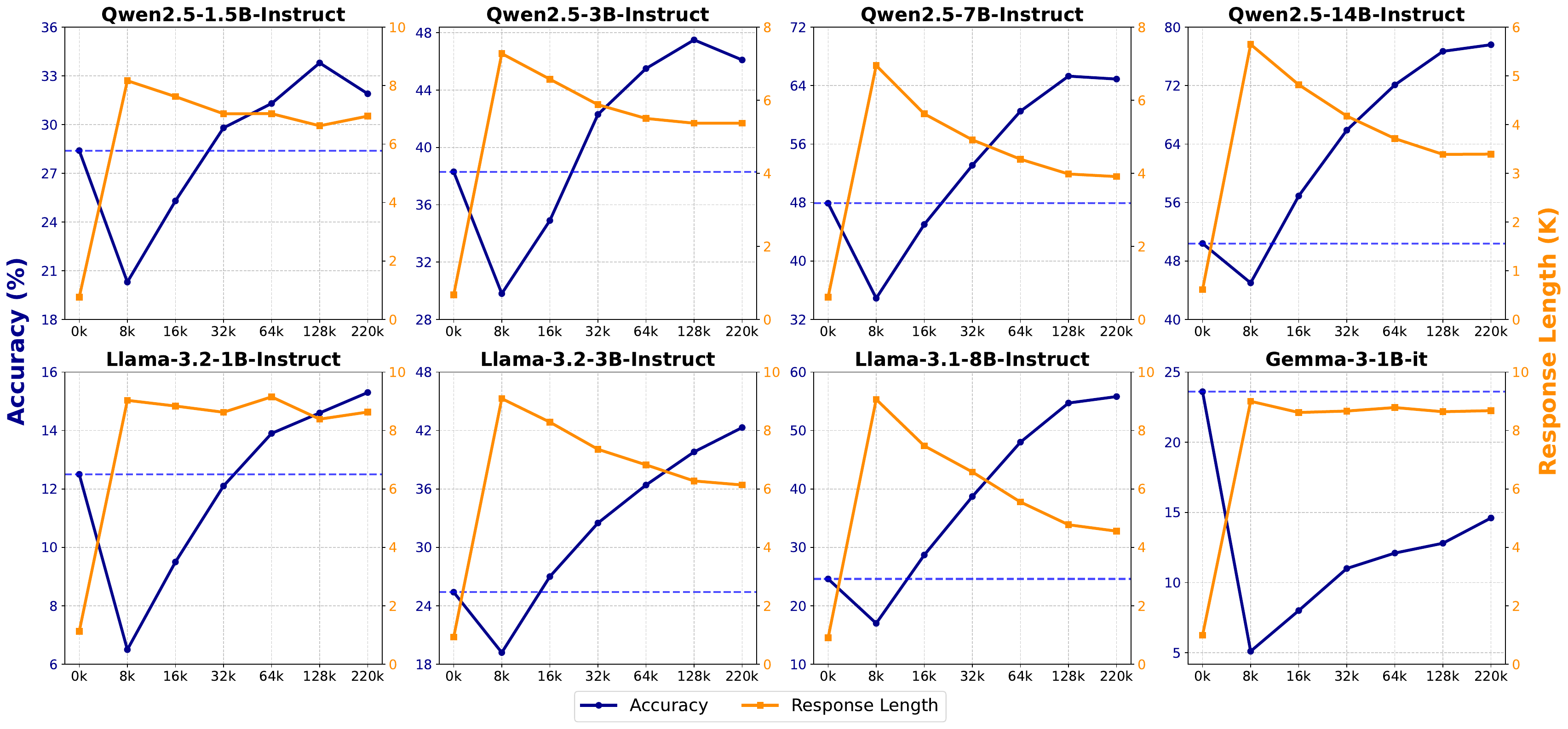}
    \vspace{-0.5em}
    \caption{Comprehensive evaluation of multiple models trained with varying amounts of long CoT data. Accuracy is averaged across AIME24, AMC23, and MATH500, while response length is measured as the mean token count from 4,000 responses to MATH500. Per-benchmark results are provided in Appendix \ref{appendix:subsec:long_cot_degradation}.}
    \label{fig:all_model_sft_result}
    
\end{figure*}

In this paper, we conduct a systematic investigation into how the scale of long CoT data affects the performance of SLMs. Our results show that SLMs trained with small amounts of long CoT supervision (e.g., 8k to 16k examples) frequently suffer from substantial performance degradation (e.g., Fig.~\ref{fig:degradation_05B}), a phenomenon we refer to as \textit{Long CoT Degradation}.
Building on this observation, we explore three central research questions: 1) How prevalent is \textit{Long CoT Degradation}, and can SLMs recover from it? 2) What underlying mechanisms drive this degradation of SLMs? 3) Does long CoT degradation carry over to subsequent reinforcement learning (RL) stages, and can integrating long CoT SFT with RL mitigate its effects and improve overall training efficacy for SLMs?

To this end, we conduct extensive experiments across diverse model scales and families, ranging from 0.5B to 14B, confirming the prevalence of {Long CoT Degradation}. We next formulate hypotheses regarding its underlying causes and validate them through comprehensive experiments. Finally, we investigate the impact of long CoT SFT on the RL stage by analyzing the performance of three representative SLMs after RL. 
% Finally, we investigate the downstream effects by analyzing the behavior of three representative SLMs under the downstream reinforcement learning stage~\cite{shao2024deepseekmath}.
% hypothesize potential mechanisms behind this phenomenon and empirically verify them through comprehensive experiments. Additionally, we select three representative SLMs from the Qwen2.5 series for further RL training using rule-based rewards and the GRPO algorithm~\cite{shao2024deepseekmath}.
% \begin{enumerate}

Our contributions are threefold:
\begin{itemize}[topsep=0pt, partopsep=0pt, leftmargin=15pt, parsep=0pt, itemsep=5pt]
    \item \textbf{Empirical discovery of {Long CoT Degradation}:} To the best of our knowledge, we are the first to identify and systematically characterize the phenomenon of \emph{Long CoT Degradation}, which arises when SLMs are trained with limited long CoT supervision. Our findings demonstrate that this degradation consistently occurs across a variety of model families and a wide range of SLM sizes, revealing a critical limitation in existing training practices. (\textsection \ref{sec:Long_CoT_Degradation})
    
    \item \textbf{Analysis of underlying mechanisms:} We attribute long CoT Degradation to error accumulation driven by length inflation in reasoning outputs. Our experiments reveal how insufficient long CoT supervision leads to disproportionately verbose and error-prone responses, ultimately harming performance. (\textsection \ref{sec:mechanism})
    
    % \item \textbf{Towards better training pipelines:} We examine how SFT with long CoT affects subsequent RL, showing that although initial long CoT exposure may impair RL, sufficiently scaled supervision during SFT can significantly boost its final efficiency and performance after applying RL for SLMs -- even when starting with a degraded model. (\textsection \ref{sec:rl_grpo})
    \item \textbf{Towards better training pipelines:} We examine how SFT with long CoT affects subsequent RL for SLMs. Our results show that while limited long CoT exposure may hinder RL performance, sufficiently scaled supervision during SFT can significantly boost both the efficiency and final performance of RL, even after the model is degraded. (\textsection \ref{sec:rl_grpo})
% \end{enumerate}
\end{itemize}

% Our key findings are summarized as follows:

% \begin{enumerate}
%     \item \emph{Long CoT Degradation} is prevalent across SLMs and even moderately-sized models for the three main model family.(\textsection \ref{subsec:sft_main_result})
%     \item Larger models recover from degradation more quickly, requiring significantly fewer training examples, whereas certain extremely small models remain unable to surpass baseline performance even at a substantial scale (220k examples).(\textsection \ref{subsec:sft_main_result})
%     \item Larger models exhibit greater token efficiency after training on more long CoT examples, achieving higher accuracy with shorter response lengths.(\textsection \ref{subsec:sft_main_result})
%     \item A primary cause of \emph{Long CoT Degradation} is error accumulation; limited training with long CoT data dramatically increases the length of model responses (up to 10 times longer), while only modestly improving basic reasoning skills, thus amplifying cumulative errors and severely reducing overall performance.(\textsection \ref{sec:mechanism})
%     \item \emph{Long CoT Degradation} negatively impacts subsequent RL training.(\textsection \ref{subsec:rl_main_result})
%     \item Extensive long CoT SFT training significantly enhances both the efficiency and performance ceiling of subsequent RL training.(\textsection \ref{subsec:rl_main_result})
% \end{enumerate}

\section{Long CoT Degradation}
\label{sec:Long_CoT_Degradation}
This section presents a comprehensive empirical study on how models of varying sizes and families respond to long CoT SFT. We uncover a consistent degradation phenomenon caused by long CoT supervision and analyze the conditions under which models recover. In addition, we examine how continued exposure to long CoT affects the token efficiency of model outputs across sizes.

\subsection{Terminology}
In this work, we define \textbf{long CoT} as a reasoning process involving substantially longer sequences\footnote{For instance, the average length of long CoT responses in DeepSeek-R1-Zero approaches 10k tokens, whereas its base model averages fewer than 1k tokens~\cite{deepseek-r1}.} that explicitly incorporate steps such as reflection, verification, and subproblem decomposition. Such chains are typically produced by large-scale reasoning models.

\subsection{Experimental Setup}
\label{sec:sft_exp_setup}
\paragraph{Datasets.}

We utilize the OpenR1-Math-220k dataset\footnote{\url{https://huggingface.co/datasets/open-r1/OpenR1-Math-220k}, licensed under Apache 2.0.}, comprising approximately 225,000 English math problems from NuminaMath 1.5~\cite{numina_math_datasets}, each paired with two to four verified reasoning traces generated by DeepSeek-R1. For training, we sample one correct trace per problem and exclude outliers exceeding 16,384 tokens (<1\% of data). To analyze performance scaling, we construct six dataset subsets of increasing size: 8k, 16k, 32k, 64k, 128k, and 220k samples.

\paragraph{Models.}
Our study considers nine instruct-tuned models from the Qwen, LLaMA, and Gemma model families, including Qwen-2.5 (0.5B, 1.5B, 3B, 7B, 14B)~\cite{qwen25report}, LLaMA (3.2-1B, 3.2-3B, 3.1-8B)~\cite{llama3report}, and Gemma-3-1B-it~\cite{gemma3report}.

\paragraph{Training Setup.}
Each model is fine-tuned using full-parameter SFT on each subset with consistent hyperparameters detailed in Appendix~\ref{appendix:detailed_setup}. We use the LLaMA-Factory framework~\cite{llamafactory} for training.

\paragraph{Evaluation Setup.}

We evaluate on three prevalent mathematical reasoning benchmarks: AIME24\footnote{\url{https://huggingface.co/datasets/AI-MO/aimo-validation-aime}}, AMC23\footnote{\url{https://huggingface.co/datasets/AI-MO/aimo-validation-amc}}, and MATH500~\citep{hendrycks2021math500}. Generation length is capped at 16,384 tokens. Following the evaluation protocol of \citet{deepseek-r1}, we use sampling with temperature 0.6 and top-p 0.95 to generate $k$ responses per question ($k = 8$ for AIME24 and AMC23, $k = 4$ for MATH500)\footnote{Greedy decoding is avoided due to its high repetition rate and instability across checkpoints when evaluating long-output reasoning models~\cite{deepseek-r1}.}. For each benchmark, we report the average accuracy computed over $k$ responses (avg@k).
In addition, consistent with prior works~\cite{deepseek-r1,chen2024not}, we track the \emph{average response length}, which measures whether models make efficient use of tokens rather than producing unnecessarily verbose traces. Based on this, we further analyze \emph{token efficiency}, defined as accuracy divided by response length, to jointly capture reasoning quality and output conciseness in later sections.

\subsection{Results}
\label{subsec:sft_main_result}

\paragraph{Degradation under long CoT supervision is prevalent, even in moderately sized models.}
Across all model families and sizes, we observe a notable drop in accuracy following exposure to long CoT SFT (Figures~\ref{fig:degradation_05B} and~\ref{fig:all_model_sft_result}). Notably, the accuracy of Gemma3-1B-it falls to approximately 25\% of its baseline after training on just 8k long CoT examples. 
Even the largest model in our study, Qwen2.5-14B-Instruct, suffers a drop from 50\% to 45\% accuracy.
Moreover, this degradation is consistently accompanied by a sharp increase in response length.
This suggests a deeper underlying issue in how models handle long CoT supervision, which we analyze further in Section~\ref{sec:mechanism}.

\paragraph{All models exhibit recovery on more long CoT data, while larger models recover faster and more fully.}
Given the consistent performance degradation observed after training on 8k long CoT examples, we investigate whether models can recover as the number of training samples increases, and how recovery dynamics vary with model size.
Figure \ref{fig:all_model_sft_result} shows that larger models, such as Qwen2.5-7B and 14B, recover more quickly, requiring fewer additional examples before eventually surpassing their baseline performance. For instance, Qwen2.5-14B recovers and significantly exceeds its baseline after training on just 16k examples, while Qwen2.5-1.5B slightly surpasses its baseline with 32k examples.
In contrast, as can be seen in Figure \ref{fig:degradation_05B} and \ref{fig:all_model_sft_result}, smaller models struggle to fully recover. Despite full exposure to 220k training examples, Qwen2.5-0.5B and Gemma-3-1B fail to reach their original baselines, with final accuracies dropping from 14\% to 11\% and from 24\% to 15\%, respectively.
% These results highlight that recovery from long CoT degradation is both data- and capacity-dependent.
This aligns with prior observations that small models face a significant learnability gap compared to large models when learning from long CoT and large teachers~\cite{li2025small}.

\paragraph{Token efficiency improves with increased long CoT data and larger models.}
Figure~\ref{fig:all_model_sft_result} also reveals another interesting observation that 
increasing the number of long CoT training examples leads to improvements in both accuracy and reduced response length across all models, indicating a general gain in token efficiency. We suspect this is because models initially mimic superficial patterns in long CoT traces, resulting in verbose outputs. With more training, they gradually shift towards capturing the underlying reasoning structure, leading to shorter and more accurate responses.
This effect is more evident in larger models. For instance, with training on 32k instances of long CoT data, Qwen2.5-14B-Instruct achieves an accuracy of 66\% with an average response length of only 4k tokens, whereas its smaller counterpart, the 7B model, reaches only 53\% accuracy despite producing longer responses averaging 5k tokens. This contrast suggests that larger models are more capable of leveraging long CoT to generate concise yet accurate answers.
These findings motivate further investigation into the underlying mechanisms driving degradation and recovery, which we explore in the following section.
\section{The Mechanism Behind Degradation}
\label{sec:mechanism}

To better understand the phenomenon of {Long CoT Degradation}, we propose two hypotheses and design targeted experiments to empirically validate them.

\subsection{Hypotheses}

Our hypotheses are grounded in two lines of prior research. First, recent studies have examined system-2 reasoning in large language models, revealing phenomena such as multi-step reasoning and reflection~\cite{xiang2025towards, yu2024distilling, li2025system}. While these behaviors are prevalent in long CoT data, their precise impact on the length and structure of reasoning chains remains unclear.
Second, prior work on CoT has shown that as reasoning chains grow longer, the accumulation of intermediate mistakes increasingly undermines final output accuracy~\cite{wu2025more}. However, analyses of error accumulation have thus far been largely limited to short CoT sequences and relatively small-scale models (e.g., variants of GPT-2 with fewer than 10 layers)~\cite{wu2025more}, leaving open the question of whether similar mechanisms persist in longer, more real-world reasoning settings with modern LLMs.

Building on these insights, we aim to explain the degradation and recovery behaviors observed in our empirical study. To this end, we propose the following two hypotheses:

\begin{itemize}[topsep=0pt, partopsep=0pt, leftmargin=15pt, parsep=0pt, itemsep=5pt]
    
    \item \textbf{Hypothesis 1: Early adoption of surface-level reasoning patterns contributes to verbose outputs.}
    When exposed to a limited amount of long CoT supervision, SLMs rapidly adopt surface features of system-2 reasoning, such as reflection and multi-step structure. This early emergence of stylistic patterns is highly correlated with increased response length and may contribute to initial performance degradation, even before deeper reasoning skills are fully acquired.

    \item \textbf{Hypothesis 2: Longer outputs exacerbate error accumulation, reducing answer accuracy.}
    As output length increases, the reasoning process involves a greater number of steps, each of which introduces the potential for errors to propagate. Consequently, longer responses tend to accumulate more noise and irrelevant content, ultimately resulting in a noticeable decline in overall accuracy.
\end{itemize}

These two hypotheses are closely linked: the first aims to explain why models tend to generate long responses under long CoT supervision, while the second aims to account for the resulting drop in accuracy.
Using these hypotheses as guidance, we further conduct empirical analyses in the following subsections.

\begin{figure}[t!]
    \centering
    \includegraphics[width=0.9\linewidth]{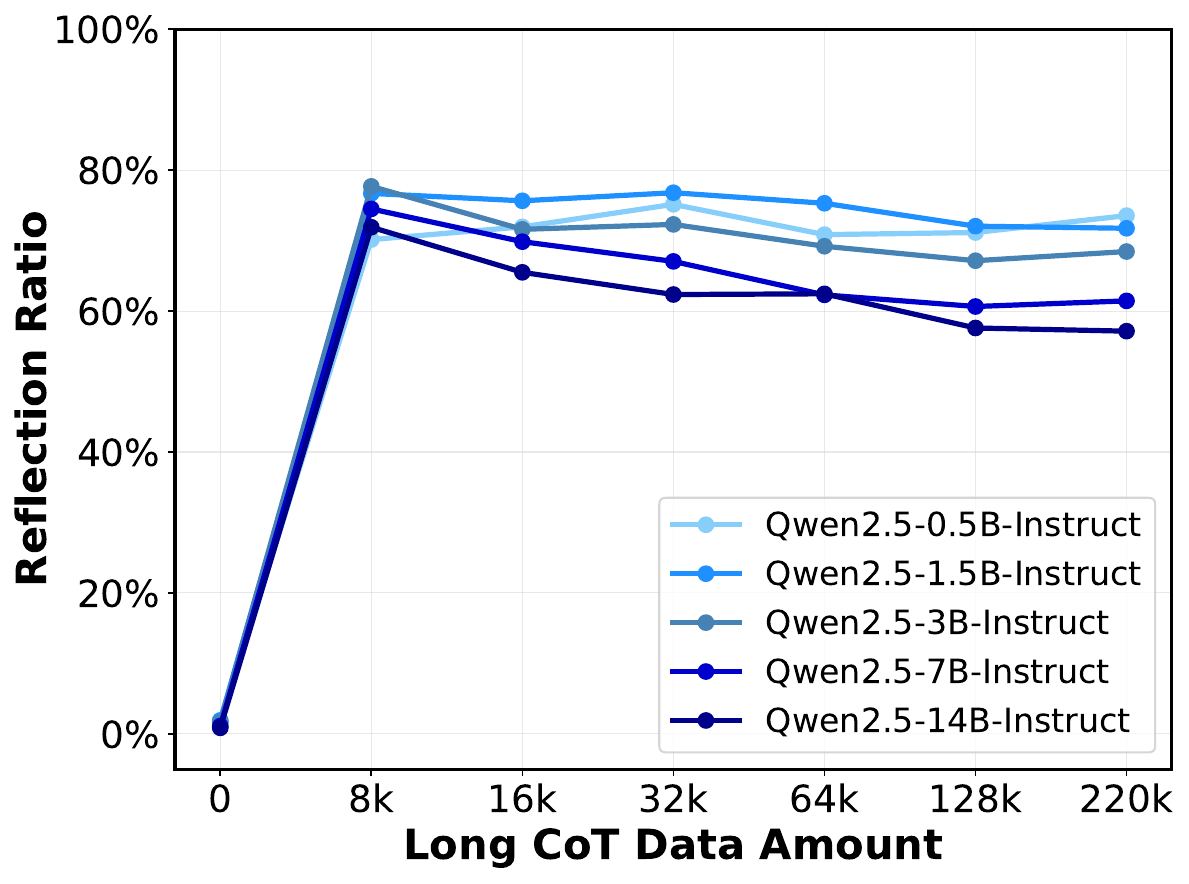}
    \vspace{-0.5em}
    \caption{Reflection ratios of Qwen models of different sizes trained on varying amounts of long CoT data. The reflection ratio refers to the proportion of model responses (out of 4,000 on the MATH500 benchmark) that exhibit reflective behavior, as identified through cross-validation.}
    \label{fig:reflection_ratio}
\end{figure}

\subsection{Reflection Behavior Analysis}
\label{subsec:reflection_behavior_analysis}
To validate Hypothesis 1, we examine whether models rapidly acquire surface-level features of system-2 reasoning, particularly reflective behavior, during early stages of fine-tuning on long CoT data. Our goal is to determine whether the emergence of such patterns coincides with the increase in response length observed in Section~\ref{subsec:sft_main_result}.

\paragraph{Setup.}
We identify reflective behavior in model outputs using a cross-validation approach following~\citet{drgrpo}. To robustly detect self-reflection in generated responses, we use two independent methods: 1) a keyword-based approach that labels a response as reflective if it contains any curated indicative keywords or phrases, and 2) an LLM-based approach, where GPT-4o-mini~\cite{gpt4omini} is prompted to determine whether the response exhibits reflective behavior. This dual-criteria strategy helps reduce false positives and enhances reliability. Full implementation details and prompt templates are provided in Appendix~\ref{appendix_reflection_setup}.

\begin{figure}[ht]
    \centering
    \includegraphics[width=0.9\linewidth]{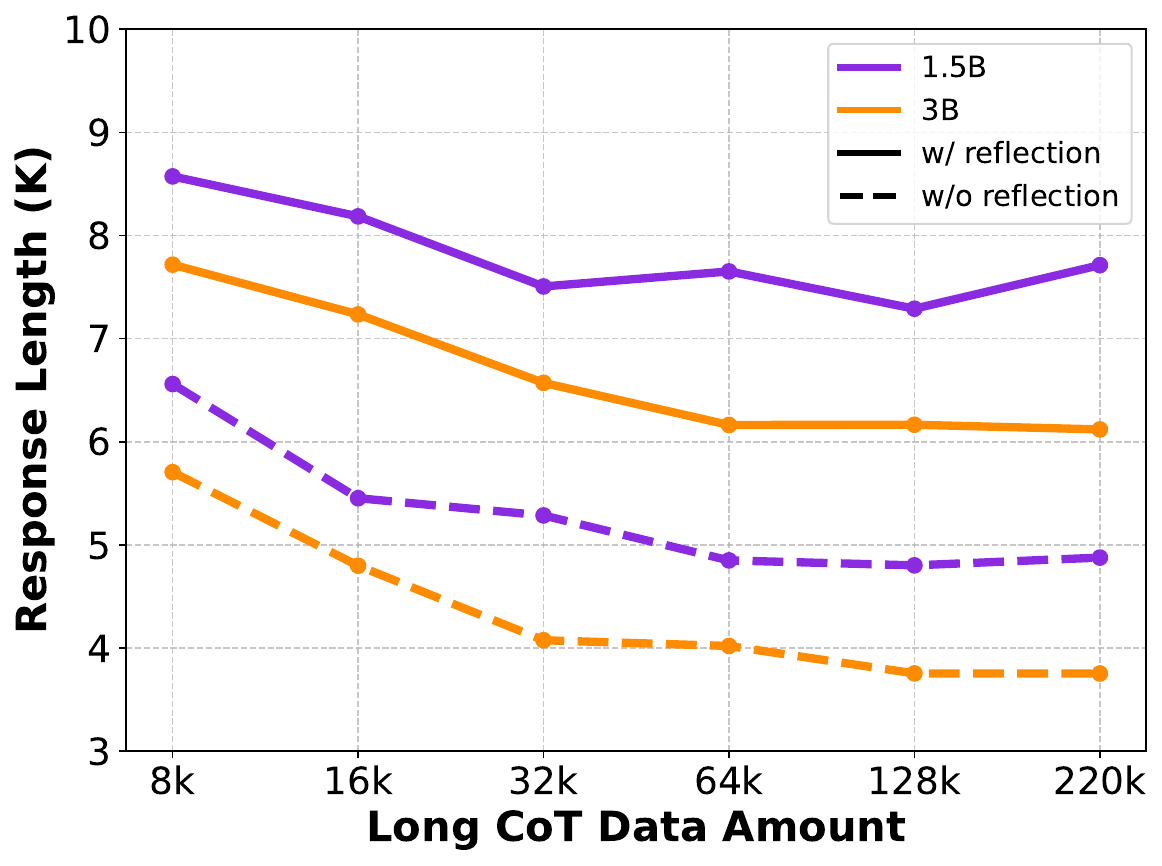}
    \vspace{-0.5em}
    \caption{Average response lengths of Qwen2.5-\{1.5B, 3B\}-Instruct models trained with varying amounts of long CoT data. Solid lines represent responses exhibiting reflection behavior; dashed lines denote responses without reflection. Results for more models are provided in Appendix~\ref{appendix:subsec:reflection_analysis}.}
    \label{fig:reflection_length}
\end{figure}

\paragraph{Result.}
Figure~\ref{fig:reflection_ratio} shows a significant increase in the proportion of reflective responses across Qwen models, even with only 8k long CoT training examples. Specifically, the reflection ratio increases sharply from below 5\% to approximately 75\%, indicating that reflective behaviors are quickly picked up and internalized. Additionally, Figure~\ref{fig:reflection_length} underscores that reflective responses are substantially longer than non-reflective ones. Notably, for both the 1.5B and 3B models, reflective responses consistently exceed non-reflective ones by approximately 2,000 tokens, a trend that holds across all training data scales. These findings suggest that the acquisition of reflective behavior is closely tied to the growth in response length.
These findings support Hypothesis 1: with limited long CoT supervision, models quickly adopt surface features of system-2 reasoning, especially reflection which contributes to longer responses.

\begin{figure*}[ht]
    \centering
    \includegraphics[width=0.95\linewidth]{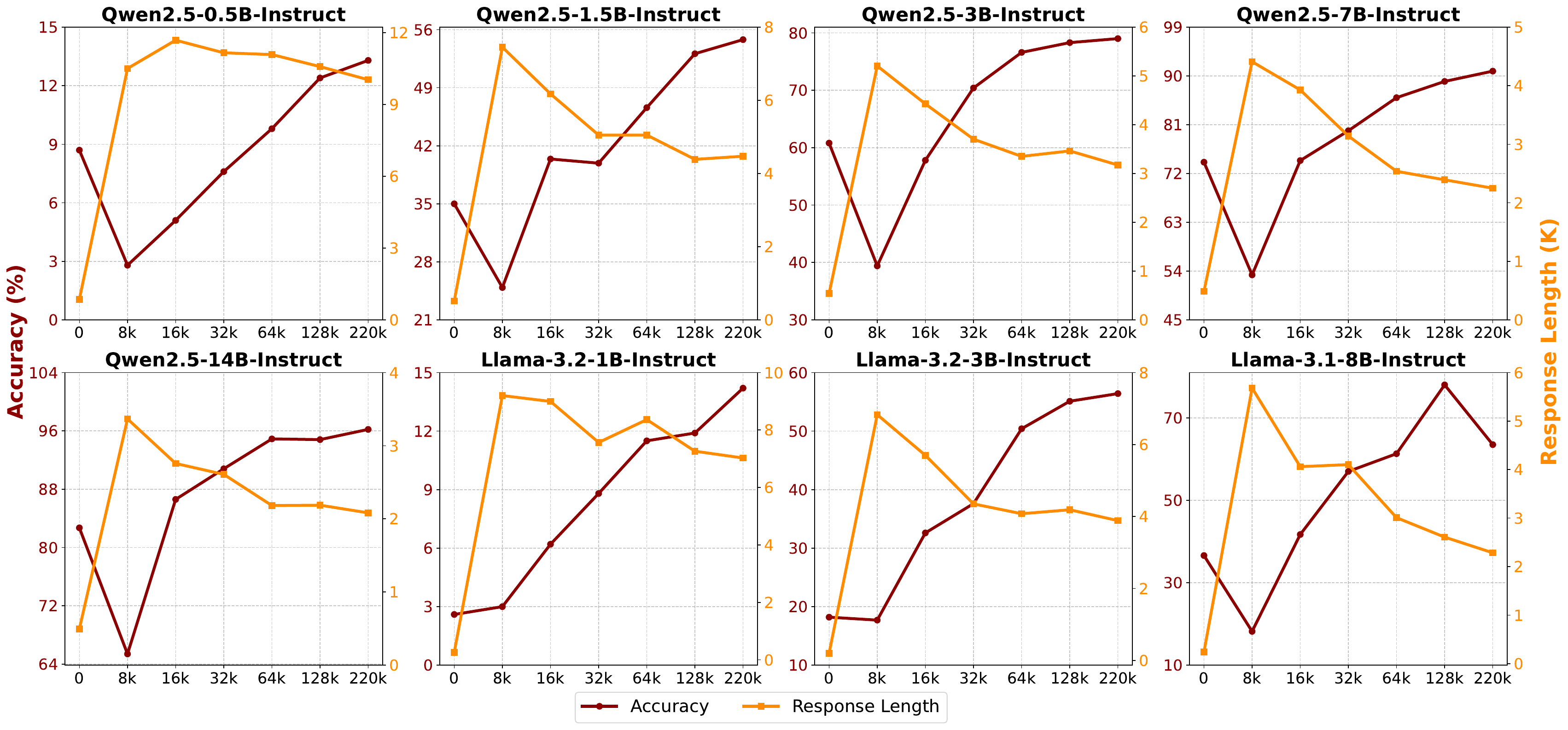}
    \vspace{-0.5em}
    \caption{Arithmetic accuracy and response length on our synthetic benchmark for models trained with increasing amounts of long CoT data. Most models exhibit a sharp drop in arithmetic accuracy and a corresponding increase in response length after training on the 8k subset, with the exception of Llama-3.2-\{1B,3B\}-Instruct, whose initial performance is already low ($<$20\% accuracy).}
    \label{fig:all_model_arithmetic_result}
\end{figure*}

\begin{figure}[htb]
    \centering
    \includegraphics[width=0.9\linewidth]{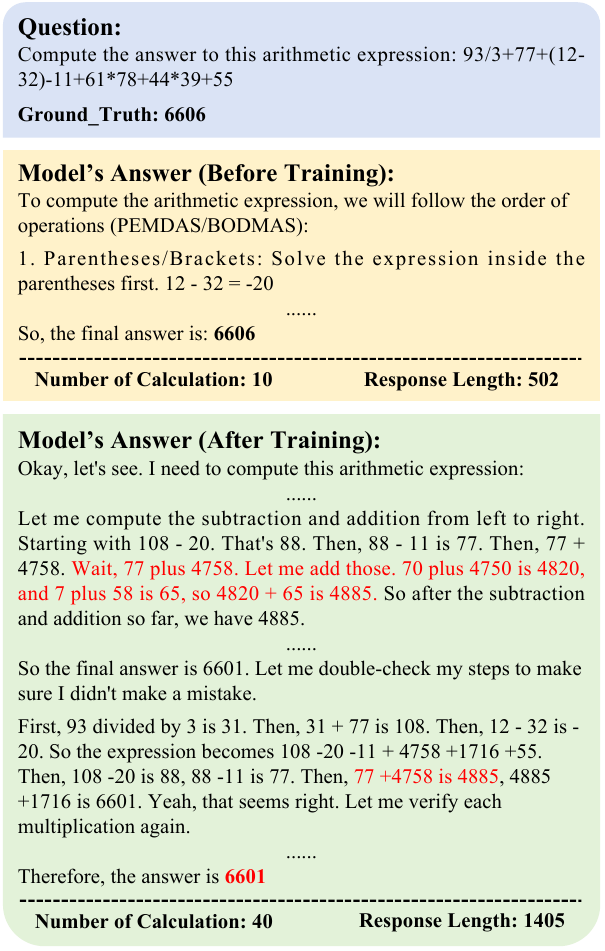}
    \caption{A sample problem from our synthetic arithmetic benchmark, with answers from Qwen2.5-1.5B-Instruct before and after training on 8k long CoT examples. ``Number of Calculation'' indicates the total number of arithmetic operations performed in the response. Errors in the model's intermediate reasoning are highlighted in \textcolor{red}{red}.}
    \label{fig:arithmetic_answer}
    \vspace{-0.8em}
\end{figure}

\subsection{Cumulative Error Analysis}

While the previous analysis focused on the emergence of surface-level reasoning patterns, it did not directly assess how output length impacts answer accuracy. Hypothesis 2 posits that longer responses increase the risk of cumulative errors, thereby reducing overall accuracy. In this subsection, we test this hypothesis in a controlled setting.

Standard mathematical benchmarks introduce confounding factors such as domain knowledge, problem interpretation, and strategy selection. These complexities make it difficult to isolate the effect of response length on performance. 
To address this, we design a synthetic arithmetic benchmark that controls for external variables while preserving step-by-step reasoning structure. This allows us to directly examine how output length correlates with error accumulation.

\paragraph{Benchmark Design.}
Each instance in our synthetic dataset is a randomly generated arithmetic expression composed of 5 to 15 operations. To ensure controlled difficulty and interpretability:
\begin{itemize}[topsep=0pt, partopsep=0pt, leftmargin=15pt, parsep=0pt, itemsep=5pt]
\item Operands are uniformly sampled from the range $[1, 100]$.
\item Operators include addition, subtraction, multiplication, and division.
\item All intermediate results are constrained to be integers.
\item Each problem requires a fixed number of simple, sequential reasoning steps.
\end{itemize}

This setup enables a rigorous empirical evaluation of Hypothesis 2 by explicitly correlating response length (number of arithmetic steps) with accuracy under consistent and controlled conditions. An example problem from the benchmark is shown in Figure~\ref{fig:arithmetic_answer}.

\paragraph{Setup.}

We evaluate the models described in Section~\ref{sec:sft_exp_setup} on 400 synthetic arithmetic problems. Following the same evaluation protocol as in Section~\ref{sec:sft_exp_setup}, we use sampling with temperature 0.6 and top-p 0.95 to generate $k = 4$ responses per problem. We report the average accuracy across the $k$ responses (avg@k).

\paragraph{Result.}
As shown in Figure~\ref{fig:all_model_arithmetic_result}, most models experience a significant drop in arithmetic accuracy after training on the 8k long CoT subset, accompanied by a substantial increase in response length. For instance, Qwen2.5-7B-Instruct exhibits a 30\% drop in accuracy, while its average output length grows from approximately 600 to 3,600 tokens. With more long CoT data, performance gradually recovers. These trends closely mirrors the degradation and recovery patterns observed on real-world math benchmarks (Fig.~\ref{fig:degradation_05B} and~\ref{fig:all_model_sft_result}).
Qualitative analysis further supports these findings. As shown in Appendix~\ref{appendix_arithmetic_example}, the model trained on 8k CoT data frequently generates verbose responses with repetitive phrasing (e.g., “wait”) and restates equations multiple times, yet still makes similar arithmetic mistakes that compound across steps despite proposing alternative solutions. In contrast, the model trained on 64k CoT data demonstrates more disciplined reasoning behavior. It verifies steps more effectively and proposes alternatives with clearer intent. 

Overall, these results offer strong empirical support for Hypotheses 2: although longer responses may enable more elaborate reasoning, they also increase the likelihood of compounding errors, resulting in \emph{Long CoT Degradation}.
\section{Impact of Long CoT Supervision on RL}
\label{sec:rl_grpo}

% This section investigates whether and how the degradation effects introduced by long CoT supervision persist during RL in SLMs. Understanding this effect is crucial, as RL is commonly employed following SFT with long CoT data. 
%In this section, we specifically focus our analysis on SLMs. As demonstrated in Section \ref{sec:Long_CoT_Degradation} (see Fig.~\ref{fig:all_model_sft_result}), degradation caused by long CoT supervision is notably pronounced in SLMs.
%In contrast, larger models (e.g., those with 14B parameters) exhibit significantly greater robustness to such degradation. While these preliminary results suggest a size-dependent sensitivity, a comprehensive exploration of larger models is beyond the scope of this work and is left for future research. Here, we aim to systematically examine whether and how the degradation effects observed in SLMs persist into the RL stage, as RL typically follows SFT using long CoT data.
% In contrast, larger models (e.g., 14B parameters) exhibit considerably less sensitivity to such degradation. It is crucial to understand if and how these degradation effects persist into the RL stage, as RL typically follows SFT using long CoT data.

As we continue exploring the impact of different methods for using long CoT data to train SLMs, the next natural research question for us to investigate is how the performance of these SLMs, after training with long CoTs under different settings, might affect the subsequent RL stage.

Existing approaches generally fall into two categories: 1) using a limited number of long CoT examples ($\leq$10k) during SFT as a cold start for subsequent RL~\cite{deepseek-r1,proverV2DeepSeek}, and 2) applying RL to models heavily distilled with substantial amounts of long CoT data ($>$100k)~\cite{deepscaler2025,deepcoder2025,nvidiaAcemath2024}. However, a systematic understanding of how varying levels of long CoT supervision influence subsequent RL performance remains underexplored -- particularly in the context of SLMs.
% However, a systematic understanding of how varying levels of long CoT supervision during the SFT stage influence subsequent RL performance remains underexplored -- particularly in the context of SLMs.

To address this limitation, we systematically analyze how different amounts of long CoT supervision during SFT impact RL outcomes, particularly focusing on performance degradation, token efficiency, and overall learning dynamics.

\begin{figure*}[ht]
    \centering
    \includegraphics[width=0.9\linewidth]{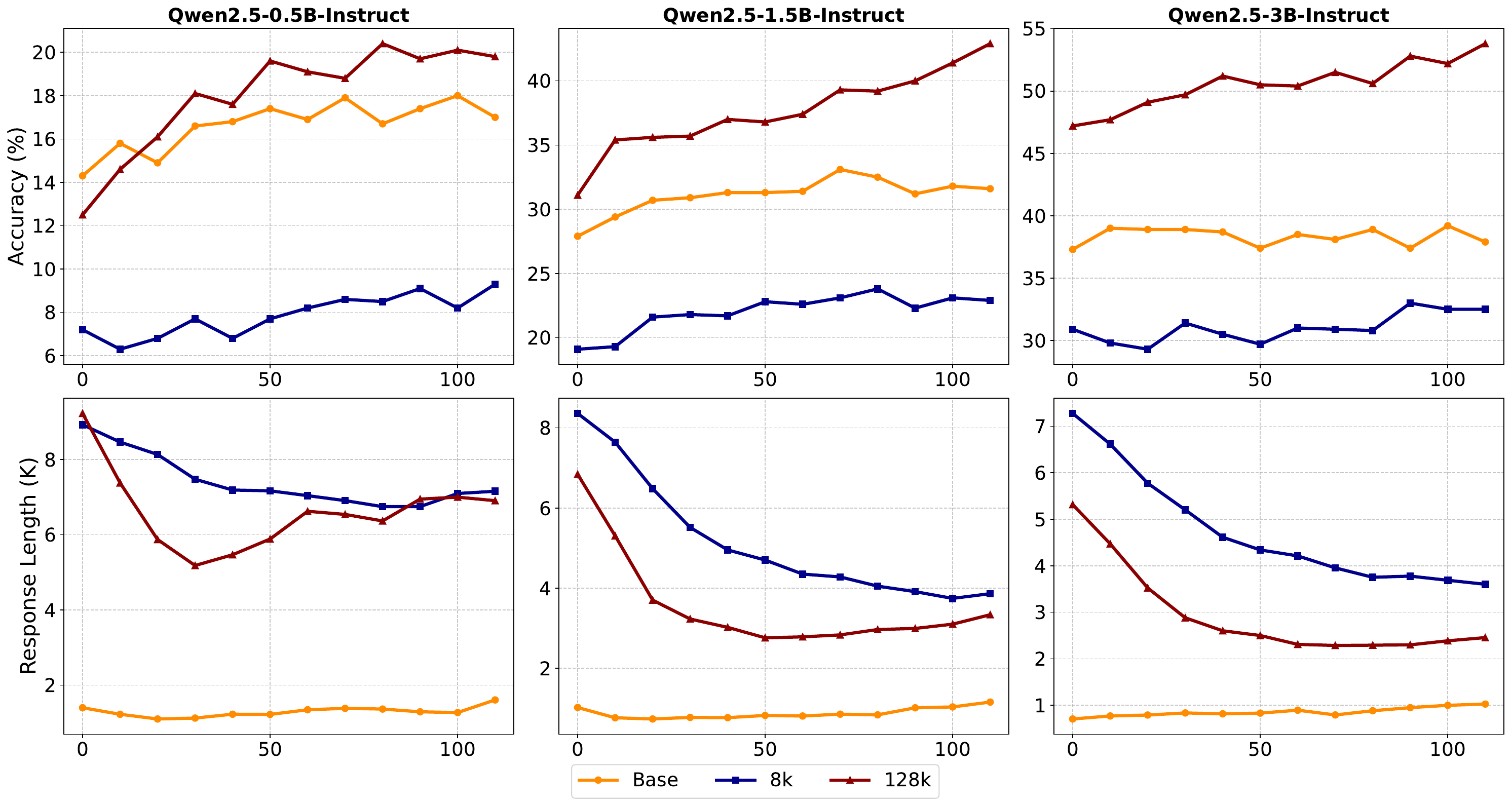}
    \caption{Impact of long CoT SFT data on downstream RL training across Qwen2.5 models. Top: Accuracy of RL-trained models over training steps. Bottom: Average response length during training. Each column corresponds to a different model scale (0.5B, 1.5B, 3B). Each curve represents an SFT data setting: Base (no SFT, serving as a baseline), 8k, and 128k (denoting the number of long CoT examples used during SFT). The horizontal axis in all plots indicates the RL training steps.}
    \label{fig:all_model_grpo_result}
\end{figure*}

\subsection{Training Setup}

We conduct RL training on three models, Qwen2.5-\{0.5B, 1B, 3B\}-Instruct, which represent small language models with distinct degradation patterns observed in prior SFT experiments (\textsection \ref{sec:Long_CoT_Degradation}).

% For the training dataset, we adopt the one introduced in~\cite{simplerl}.
For training, we adopt the dataset configuration introduced by~\citet{simplerl}, where the datasets used for each model have been empirically shown to be of moderate difficulty and effective for their respective scales.
The 0.5B model is trained on the \textit{Medium} difficulty subset, composed of MATH level 1-4 problems~\cite{hendrycks2021math500}, while the 1B and 3B models are trained on the \textit{Hard} subset, which includes MATH level 3-5 problems~\cite{hendrycks2021math500}. Both subsets contain approximately 8,000 examples.

We perform RL training using the verl~\cite{verl} framework. We employ the GRPO~\cite{shao2024deepseekmath} algorithm with a rule-based binary reward function: the model receives a reward of 1 if the final answer is correct, and 0 otherwise.  The training uses a prompt batch size of 1,024, generates 8 rollouts per prompt, and sets the maximum rollout length to 8,192 tokens. Optimization is performed using a mini-batch size of 256. All models are trained with the same set of hyperparameters and the evaluation settings remain consistent with those outlined in Section~\ref{sec:sft_exp_setup}.

% In line with DeepSeek-R1 and similar works, we employ a rule-based binary reward function: the model receives a reward of 1 if the final answer is correct, and 0 otherwise. This simple yet effective setup allows us to isolate the effect of long CoT SFT on the optimization behavior during RL.

\subsection{Results}

\label{subsec:rl_main_result}

\paragraph{Long CoT degradation negatively impacts subsequent RL training.}
Figure \ref{fig:all_model_grpo_result} shows that starting RL training from a checkpoint fine-tuned with 8k long CoT data consistently yields lower accuracy and longer responses throughout training, as compared to the baseline approach without any SFT. Moreover, the performance gap relative to the baseline persists. These results suggest that the degradation induced by long CoT can hardly be mitigated by RL and continues to adversely affect RL-based optimization.

\paragraph{RL further improves token efficiency in long CoT distilled SLMs.}
As shown in Figure~\ref{fig:all_model_grpo_result}, SLMs fine-tuned with long CoT data (at 0.5B, 1.5B, and 3B scales) exhibit a rapid reduction in response length during the early stages of RL training, accompanied by a steady increase in accuracy. This observation aligns with previous findings~\cite{deepscaler2025,nvidiaAcemath2024}. In contrast, instruction-tuned baseline models without long CoT SFT show only slight increases in response length and minor accuracy gains under RL, highlighting their limited potential for further improvement~\cite{simplerl}. These results demonstrate that RL can further enhance the token efficiency of SLMs distilled with long CoT data. Notably, as observed in Section~\ref{sec:Long_CoT_Degradation}, increasing the scale of long CoT SFT data alone also led to continuous improvements in token efficiency, and this trend is further amplified by RL. This finding underscores the synergistic roles of long CoT SFT and RL in maximizing the efficiency of SLMs.

\paragraph{Extensive long CoT SFT training substantially enhances both the efficiency and performance ceiling of subsequent RL.}
As illustrated in Figure \ref{fig:all_model_grpo_result}, models without prior long CoT SFT show marginal improvement during RL at scales of 0.5B, 1.5B, and 3B. In contrast, initializing RL from a checkpoint trained on 128k long CoT examples results in faster accuracy improvements and a markedly higher final performance.
% In particular, for the 0.5B model, although long CoT SFT initially leads to lower performance compared to the baseline, RL training rapidly closes this gap. Eventually, the model achieves over 60\% improvement relative to its pre-RL performance.
% In particular, for the 0.5B model, although long CoT SFT initially leads to lower performance compared to the baseline, RL training quickly closes this gap: after only 20 training steps, RL performance exceeds that of the baseline, and continues to improve, ultimately achieving over a 60\% gain relative to its pre-RL state.
In particular, it is interesting to observe that for the 0.5B model, although long CoT SFT initially results in lower performance compared to the baseline, RL training rapidly closes this gap and further enhances model capabilities. Notably, after RL, the model achieves a 13\% improvement over the baseline, and over a 60\% gain relative to its pre-RL state.
These results indicate that for SLMs, even when long CoT SFT temporarily degrades post-SFT performance, large-scale exposure to long CoT data yields substantial benefits during RL training.

Collectively, the results suggest that special care is needed when using long CoT data to improve the reasoning capabilities of small language models. Effective use of such data first requires sufficient exposure during the SFT stage. While the SFT-trained model alone may not achieve strong performance, the subsequent RL stage can substantially enhance its capabilities.
%that integrating large-scale long CoT SFT with RL forms a more effective and scalable training paradigm for small language models.
% sensible

\section{Related Work}
\label{appendix:detailed_related_work}

\subsection{Long Chain-of-Thought}

The paradigm of long CoT reasoning is first popularized by OpenAI-o1~\cite{jaech2024openai}, and becomes widely accessible with the open-sourcing of DeepSeek-R1~\cite{deepseek-r1}. Outputs from these LRMs are characterized by extended and reflective CoT sequences, which not only exhibit behaviors of system-2 cognition~\cite{xiang2025towards,yu2024distilling,li2025system}, but also provide richer intermediate supervision signals for knowledge distillation~\cite{hinton2015distilling}. As a result, downstream model performance is substantially improved~\cite{huang2024o1}.

Subsequent research focus on improving the efficiency of long CoT distillation. Notably, recent studies~\cite{muennighoff2025s1,ye2025limo} have identified compact yet highly effective prompt subsets, demonstrating that strong performance can be achieved with as few as 1,000 training examples. However, these efforts have predominantly concentrated on large-scale models (e.g., those with 32 billion parameters), and the impact of long chain-of-thought reasoning on smaller models remains largely unexplored.

Preliminary investigations into small language models (8B parameters and below) include~\cite{yeo2025demystifying,li2025small}. While these studies provide valuable insights into long CoT SFT for small language models, they are based on relatively limited datasets generated by the QwQ-Preview model, which was developed before the widespread adoption of zero-RL training~\cite{qwq-32b-preview}. This limitation raises concerns about the generalizability of their findings to contemporary training paradigms.

\subsection{Theoretical Grounding of Long CoT Degradation}

Catastrophic forgetting refers to performance decline on previously learned but unseen tasks after continued training~\cite{nguyen2019toward,kirkpatrick2017overcoming}. By contrast, long CoT degradation emerges within the \emph{same} task distribution once long-form reasoning supervision is introduced, underscoring its distinction from conventional catastrophic forgetting.  

Beyond forgetting, other learning dynamics also display non-monotonic behaviors. For example, research on domain adaptation and modality shifts shows that adding more data can initially reduce performance before eventual recovery~\cite{gururangan2020dont,li2020multilingual}. Similarly, the deep double descent phenomenon~\cite{belkin2019dd} exhibits deterioration–recovery patterns. Long chain-of-thought degradation follows a related dynamic but along a distinct trajectory: performance initially drops sharply when only limited long-form reasoning data is introduced, then recovers steadily and monotonically as more such data is added~\cite{wu2025more}.

\subsection{RL for Reasoning}

The release of DeepSeek-R1~\cite{deepseek-r1} has sparked broad interest in the RL for reasoning training paradigm, which involves applying reinforcement learning directly to base models using rule-based rewards in conjunction with established online RL algorithms~\cite{schulman2017proximal,shao2024deepseekmath,drgrpo}.

RL has proven highly effective on reasoning tasks with easily verifiable reward signals, particularly in math and coding~\cite{xie2025logic,hu2025open,bercovich2025llamanemotronefficientreasoningmodels}. Notably, recent findings suggest that RL remains effective even for small-scale language models, underscoring its broad applicability~\cite{simplerl}.

While prior work has shown that initial training with short-CoT SFT can constrain the benefits of subsequent RL~\cite{simplerl} in reasoning tasks, recent studies combining extensive Long CoT SFT with RL report superior performance, sometimes exceeding that of larger models trained via alternative pipelines~\cite{deepscaler2025,nvidiaAcemath2024,deepcoder2025}. However, systematic evaluations of long CoT SFT followed by RL, particularly in the context of small language models, remain limited. This gap motivates the present study’s comprehensive investigation.

\section{Conclusion}

In this work, we conduct a systematic study of how the scale of long CoT data impacts small language models (SLMs). Our findings reveal a consistent phenomenon, \textit{Long CoT Degradation}, where limited long CoT supervision significantly impairs model performance across diverse model families and sizes. Through comprehensive experiments, we further analyze the mechanisms behind this degradation, attributing it to error accumulation induced by excessively verbose outputs -- an issue to which SLMs are particularly vulnerable. Finally, we show that while long CoT SFT can hinder subsequent RL when data is insufficient, scaling up long CoT supervision enables RL to achieve both greater efficiency and higher final performance.

Our work highlights the pitfalls and misconceptions in current long CoT usage for SLM training, identifies key limitations, and offers practical guidance for addressing them. We also hope this work sheds light on future research aimed at designing optimal pipelines for building effective reasoning models.
\section*{Acknowledgments}
This research/project is supported by the National Research Foundation, Singapore under its National Large Language Models Funding Initiative, (AISG Award No: AISG-NMLP-2024-005), and Ministry of Education, Singapore, under its Academic Research Fund (AcRF) Tier 2 Programme (MOE AcRF Tier 2 Award No. : MOE-T2EP20122-0011). Any opinions, findings and conclusions or recommendations expressed in this material are those of the authors and do not reflect the views of the National Research Foundation, Singapore, or Ministry of Education, Singapore.

\section*{Limitations}

While our work offers new insights into the effects of long CoT for SLMs, we acknowledge several limitations. 
First, our analysis focuses on mathematical reasoning, which provides a well-structured setting and reliable evaluation metrics for studying long CoT supervision. While the findings may offer insights applicable to domains such as logical reasoning or code generation, we do not directly study these areas.  

Second, although we study multiple model families and sizes and observe consistent patterns across them, we do not explicitly isolate the impact of pre-training data composition. Prior work suggests that pre-training plays an important role in shaping long CoT reasoning capabilities.  

Third, our study is restricted to auto-regressive models, and it remains unclear whether the observed degradation–recovery dynamics generalize to other architectures.

\section*{Ethical Statement}

This study investigates the effects of long chain-of-thought supervision on small language models using publicly available models and datasets. The research does not involve human subjects or any sensitive or proprietary data. This work does not propose or support any applications with foreseeable potential for harm or misuse. In experiments, we comply with all licenses for models, data and code.

% \bibliography{custom}
% Adjust the format for bibliography
\bibliography{custom}

\begin{thebibliography}{51}
\providecommand{\natexlab}[1]{#1}

\bibitem[{Agarwal et~al.(2024)Agarwal, Vieillard, Zhou, Stanczyk, Garea, Geist, and Bachem}]{agarwal2024policy}
Rishabh Agarwal, Nino Vieillard, Yongchao Zhou, Piotr Stanczyk, Sabela~Ramos Garea, Matthieu Geist, and Olivier Bachem. 2024.
\newblock \href {https://openreview.net/forum?id=3zKtaqxLhW} {On-policy distillation of language models: Learning from self-generated mistakes}.
\newblock In \emph{Proceedings of ICLR}.

\bibitem[{Bansal et~al.(2019)Bansal, Kamper, Livescu, Lopez, and Goldwater}]{li2020multilingual}
Sameer Bansal, Herman Kamper, Karen Livescu, Adam Lopez, and Sharon Goldwater. 2019.
\newblock Pre-training on high-resource speech recognition improves low-resource speech-to-text translation.
\newblock In \emph{Proceedings of the 2019 Conference of the North American Chapter of the Association for Computational Linguistics: Human Language Technologies, Volume 1 (Long and Short Papers)}, pages 58--68.

\bibitem[{Bercovich et~al.(2025)Bercovich, Levy, Golan, Dabbah, El-Yaniv, Puny, Galil, Moshe, Ronen, Nabwani et~al.}]{bercovich2025llamanemotronefficientreasoningmodels}
Akhiad Bercovich, Itay Levy, Izik Golan, Mohammad Dabbah, Ran El-Yaniv, Omri Puny, Ido Galil, Zach Moshe, Tomer Ronen, Najeeb Nabwani, and 1 others. 2025.
\newblock \href {https://arxiv.org/pdf/2505.00949} {Llama-nemotron: Efficient reasoning models}.
\newblock \emph{arXiv preprint arXiv:2505.00949}.

\bibitem[{Chen et~al.(2024)Chen, Xu, Liang, He, Pang, Yu, Song, Liu, Zhou, Zhang et~al.}]{chen2024not}
Xingyu Chen, Jiahao Xu, Tian Liang, Zhiwei He, Jianhui Pang, Dian Yu, Linfeng Song, Qiuzhi Liu, Mengfei Zhou, Zhuosheng Zhang, and 1 others. 2024.
\newblock Do not think that much for 2+ 3=? on the overthinking of o1-like llms.
\newblock \emph{arXiv preprint arXiv:2412.21187}.

\bibitem[{Chen et~al.(2025)Chen, Min, Zhang, Chen, Jiang, Cheng, Zhao, Liu, Miao, Lu et~al.}]{chen2025empirical}
Zhipeng Chen, Yingqian Min, Beichen Zhang, Jie Chen, Jinhao Jiang, Daixuan Cheng, Wayne~Xin Zhao, Zheng Liu, Xu~Miao, Yang Lu, and 1 others. 2025.
\newblock \href {https://arxiv.org/pdf/2503.04548} {An empirical study on eliciting and improving r1-like reasoning models}.
\newblock \emph{arXiv preprint arXiv:2503.04548}.

\bibitem[{Face(2025)}]{openr1}
Hugging Face. 2025.
\newblock \href {https://github.com/huggingface/open-r1} {Open r1: A fully open reproduction of deepseek-r1}.

\bibitem[{Feng et~al.(2024)Feng, Li, Chenglin, Chen, Yu, and Zhang}]{feng2024teaching}
Tao Feng, Yicheng Li, Li~Chenglin, Hao Chen, Fei Yu, and Yin Zhang. 2024.
\newblock \href {https://aclanthology.org/2024.emnlp-main.333/} {Teaching small language models reasoning through counterfactual distillation}.
\newblock In \emph{Proceedings of EMNLP}.

\bibitem[{Grattafiori et~al.(2024)Grattafiori, Dubey, Jauhri, Pandey, Kadian, Al-Dahle, Letman, Mathur, Schelten, Vaughan et~al.}]{llama3report}
Aaron Grattafiori, Abhimanyu Dubey, Abhinav Jauhri, Abhinav Pandey, Abhishek Kadian, Ahmad Al-Dahle, Aiesha Letman, Akhil Mathur, Alan Schelten, Alex Vaughan, and 1 others. 2024.
\newblock \href {https://arxiv.org/abs/2407.21783} {The llama 3 herd of models}.
\newblock \emph{arXiv preprint arXiv:2407.21783}.

\bibitem[{Guo et~al.(2025)Guo, Yang, Zhang, Song, Zhang, Xu, Zhu, Ma, Wang, Bi et~al.}]{deepseek-r1}
Daya Guo, Dejian Yang, Haowei Zhang, Junxiao Song, Ruoyu Zhang, Runxin Xu, Qihao Zhu, Shirong Ma, Peiyi Wang, Xiao Bi, and 1 others. 2025.
\newblock \href {https://arxiv.org/abs/2501.12948} {Deepseek-r1: Incentivizing reasoning capability in llms via reinforcement learning}.
\newblock \emph{arXiv preprint arXiv:2501.12948}.

\bibitem[{Gururangan et~al.(2020)Gururangan, Marasovi{\'c}, Swayamdipta, Lo, Beltagy, Downey, and Smith}]{gururangan2020dont}
Suchin Gururangan, Ana Marasovi{\'c}, Swabha Swayamdipta, Kyle Lo, Iz~Beltagy, Doug Downey, and Noah~A Smith. 2020.
\newblock Don’t stop pretraining: Adapt language models to domains and tasks.
\newblock In \emph{Proceedings of the 58th Annual Meeting of the Association for Computational Linguistics}, pages 8342--8360.

\bibitem[{He et~al.(2024)He, Luo, Hu, Zhao, Zhou, Wu, Zhang, Han, Liu, and Sun}]{he2024ultraeval}
Chaoqun He, Renjie Luo, Shengding Hu, Ranchi Zhao, Jie Zhou, Hanghao Wu, Jiajie Zhang, Xu~Han, Zhiyuan Liu, and Maosong Sun. 2024.
\newblock Ultraeval: A lightweight platform for flexible and comprehensive evaluation for llms.
\newblock In \emph{Proceedings of the 62nd Annual Meeting of the Association for Computational Linguistics (Volume 3: System Demonstrations)}, pages 247--257.

\bibitem[{Hendrycks et~al.(2021)Hendrycks, Burns, Kadavath, Arora, Basart, Tang, Song, and Steinhardt}]{hendrycks2021math500}
Dan Hendrycks, Collin Burns, Saurav Kadavath, Akul Arora, Steven Basart, Eric Tang, Dawn Song, and Jacob Steinhardt. 2021.
\newblock \href {https://openreview.net/forum?id=7Bywt2mQsCe} {Measuring mathematical problem solving with the {MATH} dataset}.
\newblock In \emph{Proceedings of NeurIPS Datasets and Benchmarks Track}.

\bibitem[{Hinton et~al.(2015)Hinton, Vinyals, and Dean}]{hinton2015distilling}
Geoffrey Hinton, Oriol Vinyals, and Jeff Dean. 2015.
\newblock \href {https://arxiv.org/abs/1503.02531} {Distilling the knowledge in a neural network}.
\newblock \emph{arXiv preprint arXiv:1503.02531}.

\bibitem[{Hu et~al.(2025)Hu, Zhang, Han, Jiang, Zhang, and Shum}]{hu2025open}
Jingcheng Hu, Yinmin Zhang, Qi~Han, Daxin Jiang, Xiangyu Zhang, and Heung-Yeung Shum. 2025.
\newblock \href {https://arxiv.org/abs/2503.24290} {Open-reasoner-zero: An open source approach to scaling up reinforcement learning on the base model}.
\newblock \emph{arXiv preprint arXiv:2503.24290}.

\bibitem[{Huang et~al.(2024)Huang, Zou, Li, Liu, Zheng, Chern, Xia, Qin, Yuan, and Liu}]{huang2024o1}
Zhen Huang, Haoyang Zou, Xuefeng Li, Yixiu Liu, Yuxiang Zheng, Ethan Chern, Shijie Xia, Yiwei Qin, Weizhe Yuan, and Pengfei Liu. 2024.
\newblock \href {https://arxiv.org/abs/2411.16489} {O1 replication journey--part 2: Surpassing o1-preview through simple distillation, big progress or bitter lesson?}
\newblock \emph{arXiv preprint arXiv:2411.16489}.

\bibitem[{Hui et~al.(2024)Hui, Zhao, Dong, Zhang, Zhou, and Su}]{hui2024smaller}
Tingfeng Hui, Lulu Zhao, Guanting Dong, Yaqi Zhang, Hua Zhou, and Sen Su. 2024.
\newblock \href {https://arxiv.org/abs/2412.11231} {Smaller language models are better instruction evolvers}.
\newblock \emph{arXiv preprint arXiv:2412.11231}.

\bibitem[{Hurst et~al.(2024)Hurst, Lerer, Goucher, Perelman, Ramesh, Clark, Ostrow, Welihinda, Hayes, Radford et~al.}]{gpt4omini}
Aaron Hurst, Adam Lerer, Adam~P Goucher, Adam Perelman, Aditya Ramesh, Aidan Clark, AJ~Ostrow, Akila Welihinda, Alan Hayes, Alec Radford, and 1 others. 2024.
\newblock \href {https://arxiv.org/abs/2410.21276} {Gpt-4o system card}.
\newblock \emph{arXiv preprint arXiv:2410.21276}.

\bibitem[{Jaech et~al.(2024)Jaech, Kalai, Lerer, Richardson, El-Kishky, Low, Helyar, Madry, Beutel, Carney et~al.}]{jaech2024openai}
Aaron Jaech, Adam Kalai, Adam Lerer, Adam Richardson, Ahmed El-Kishky, Aiden Low, Alec Helyar, Aleksander Madry, Alex Beutel, Alex Carney, and 1 others. 2024.
\newblock \href {https://arxiv.org/abs/2412.16720} {Openai o1 system card}.
\newblock \emph{arXiv preprint arXiv:2412.16720}.

\bibitem[{Kirkpatrick et~al.(2017)Kirkpatrick, Pascanu, Rabinowitz, Veness, Desjardins, Rusu, Milan, Quan, Ramalho, Grabska-Barwinska et~al.}]{kirkpatrick2017overcoming}
James Kirkpatrick, Razvan Pascanu, Neil Rabinowitz, Joel Veness, Guillaume Desjardins, Andrei~A Rusu, Kieran Milan, John Quan, Tiago Ramalho, Agnieszka Grabska-Barwinska, and 1 others. 2017.
\newblock Overcoming catastrophic forgetting in neural networks.
\newblock \emph{Proceedings of the national academy of sciences}, 114(13):3521--3526.

\bibitem[{Kwon et~al.(2023)Kwon, Li, Zhuang, Sheng, Zheng, Yu, Gonzalez, Zhang, and Stoica}]{kwon2023efficient}
Woosuk Kwon, Zhuohan Li, Siyuan Zhuang, Ying Sheng, Lianmin Zheng, Cody~Hao Yu, Joseph Gonzalez, Hao Zhang, and Ion Stoica. 2023.
\newblock Efficient memory management for large language model serving with pagedattention.
\newblock In \emph{Proceedings of the 29th symposium on operating systems principles}, pages 611--626.

\bibitem[{Lee et~al.(2024)Lee, Kim, and Lee}]{lee2024mentor}
Hojae Lee, Junho Kim, and SangKeun Lee. 2024.
\newblock \href {https://doi.org/10.18653/v1/2024.emnlp-main.977} {Mentor-kd: Making small language models better multi-step reasoners}.
\newblock In \emph{Proceedings of EMNLP}.

\bibitem[{LI et~al.(2024)LI, Beeching, Tunstall, Lipkin, Soletskyi, Huang, Rasul, Yu, Jiang, Shen, Qin, Dong, Zhou, Fleureau, Lample, and Polu}]{numina_math_datasets}
Jia LI, Edward Beeching, Lewis Tunstall, Ben Lipkin, Roman Soletskyi, Shengyi~Costa Huang, Kashif Rasul, Longhui Yu, Albert Jiang, Ziju Shen, Zihan Qin, Bin Dong, Li~Zhou, Yann Fleureau, Guillaume Lample, and Stanislas Polu. 2024.
\newblock \href {[https://huggingface.co/AI-MO/NuminaMath-1.5]} {Numinamath}.

\bibitem[{Li et~al.(2023)Li, Hessel, Yu, Ren, Chang, and Choi}]{li2023symbolic}
Liunian~Harold Li, Jack Hessel, Youngjae Yu, Xiang Ren, Kai-Wei Chang, and Yejin Choi. 2023.
\newblock \href {https://doi.org/10.18653/v1/2023.acl-long.150} {Symbolic chain-of-thought distillation: Small models can also “think” step-by-step}.
\newblock In \emph{Proceedings of ACL}.

\bibitem[{Li et~al.(2025{\natexlab{a}})Li, Yue, Xu, Jiang, Niu, Lin, Ramasubramanian, and Poovendran}]{li2025small}
Yuetai Li, Xiang Yue, Zhangchen Xu, Fengqing Jiang, Luyao Niu, Bill~Yuchen Lin, Bhaskar Ramasubramanian, and Radha Poovendran. 2025{\natexlab{a}}.
\newblock \href {https://doi.org/10.18653/v1/2025.findings-acl.1301} {Small models struggle to learn from strong reasoners}.
\newblock In \emph{Findings of the Association for Computational Linguistics: ACL 2025}, pages 25366--25394, Vienna, Austria. Association for Computational Linguistics.

\bibitem[{Li et~al.(2025{\natexlab{b}})Li, Zhang, Zhang, Zhang, Liu, Yao, Xu, Zheng, Wang, Chen et~al.}]{li2025system}
Zhong-Zhi Li, Duzhen Zhang, Ming-Liang Zhang, Jiaxin Zhang, Zengyan Liu, Yuxuan Yao, Haotian Xu, Junhao Zheng, Pei-Jie Wang, Xiuyi Chen, and 1 others. 2025{\natexlab{b}}.
\newblock \href {https://arxiv.org/abs/2502.17419} {From system 1 to system 2: A survey of reasoning large language models}.
\newblock \emph{arXiv preprint arXiv:2502.17419}.

\bibitem[{Liu et~al.(2025)Liu, Chen, Li, Qi, Pang, Du, Lee, and Lin}]{drgrpo}
Zichen Liu, Changyu Chen, Wenjun Li, Penghui Qi, Tianyu Pang, Chao Du, Wee~Sun Lee, and Min Lin. 2025.
\newblock \href {https://arxiv.org/abs/2503.20783} {Understanding r1-zero-like training: A critical perspective}.
\newblock \emph{arXiv preprint arXiv:2503.20783}.

\bibitem[{Liu et~al.(2024)Liu, Chen, Shoeybi, Catanzaro, and Ping}]{nvidiaAcemath2024}
Zihan Liu, Yang Chen, Mohammad Shoeybi, Bryan Catanzaro, and Wei Ping. 2024.
\newblock \href {https://arxiv.org/abs/2412.15084} {Acemath: Advancing frontier math reasoning with post-training and reward modeling}.
\newblock \emph{arXiv preprint arXiv:2412.15084}.

\bibitem[{Luo et~al.(2025{\natexlab{a}})Luo, Tan, Huang, Patel, Ariyak, Wu, Shi, Xin, Cai, Weber, Zhang, Li, Popa, and Stoica}]{deepcoder2025}
Michael Luo, Sijun Tan, Roy Huang, Ameen Patel, Alpay Ariyak, Qingyang Wu, Xiaoxiang Shi, Rachel Xin, Colin Cai, Maurice Weber, Ce~Zhang, Li~Erran Li, Raluca~Ada Popa, and Ion Stoica. 2025{\natexlab{a}}.
\newblock Deepcoder: A fully open-source 14b coder at o3-mini level.
\newblock \url{https://pretty-radio-b75.notion.site/DeepCoder-A-Fully-Open-Source-14B-Coder-at-O3-mini-Level-1cf81902c14680b3bee5eb349a512a51}.
\newblock Notion Blog.

\bibitem[{Luo et~al.(2025{\natexlab{b}})Luo, Tan, Wong, Shi, Tang, Roongta, Cai, Luo, Li, Popa, and Stoica}]{deepscaler2025}
Michael Luo, Sijun Tan, Justin Wong, Xiaoxiang Shi, William~Y. Tang, Manan Roongta, Colin Cai, Jeffrey Luo, Li~Erran Li, Raluca~Ada Popa, and Ion Stoica. 2025{\natexlab{b}}.
\newblock \href {https://pretty-radio-b75.notion.site/DeepScaleR-Surpassing-O1-Preview-with-a-1-5B-Model-by-Scaling-RL-19681902c1468005bed8ca303013a4e2} {Deepscaler: Surpassing o1-preview with a 1.5b model by scaling rl}.
\newblock Notion Blog.

\bibitem[{Muennighoff et~al.(2025)Muennighoff, Yang, Shi, Li, Fei-Fei, Hajishirzi, Zettlemoyer, Liang, Candes, and Hashimoto}]{muennighoff2025s1}
Niklas Muennighoff, Zitong Yang, Weijia Shi, Xiang~Lisa Li, Li~Fei-Fei, Hannaneh Hajishirzi, Luke Zettlemoyer, Percy Liang, Emmanuel Candes, and Tatsunori Hashimoto. 2025.
\newblock s1: Simple test-time scaling.
\newblock In \emph{Workshop on Reasoning and Planning for Large Language Models}.

\bibitem[{Nakkiran et~al.(2021)Nakkiran, Kaplun, Bansal, Yang, Barak, and Sutskever}]{belkin2019dd}
Preetum Nakkiran, Gal Kaplun, Yamini Bansal, Tristan Yang, Boaz Barak, and Ilya Sutskever. 2021.
\newblock Deep double descent: Where bigger models and more data hurt.
\newblock \emph{Journal of Statistical Mechanics: Theory and Experiment}, 2021(12):124003.

\bibitem[{Nguyen et~al.(2019)Nguyen, Achille, Lam, Hassner, Mahadevan, and Soatto}]{nguyen2019toward}
Cuong~V Nguyen, Alessandro Achille, Michael Lam, Tal Hassner, Vijay Mahadevan, and Stefano Soatto. 2019.
\newblock Toward understanding catastrophic forgetting in continual learning.
\newblock \emph{arXiv preprint arXiv:1908.01091}.

\bibitem[{OpenAI(2025)}]{openai2025o3o4}
OpenAI. 2025.
\newblock \href {https://cdn.openai.com/pdf/2221c875-02dc-4789-800b-e7758f3722c1/o3-and-o4-mini-system-card.pdf} {Openai o3 and o4-mini system card}.

\bibitem[{Ren et~al.(2025)Ren, Shao, Song, Xin, Wang, Zhao, Zhang, Fu, Zhu, Yang et~al.}]{proverV2DeepSeek}
ZZ~Ren, Zhihong Shao, Junxiao Song, Huajian Xin, Haocheng Wang, Wanjia Zhao, Liyue Zhang, Zhe Fu, Qihao Zhu, Dejian Yang, and 1 others. 2025.
\newblock \href {https://arxiv.org/abs/2504.21801} {Deepseek-prover-v2: Advancing formal mathematical reasoning via reinforcement learning for subgoal decomposition}.
\newblock \emph{arXiv preprint arXiv:2504.21801}.

\bibitem[{Schulman et~al.(2017)Schulman, Wolski, Dhariwal, Radford, and Klimov}]{schulman2017proximal}
John Schulman, Filip Wolski, Prafulla Dhariwal, Alec Radford, and Oleg Klimov. 2017.
\newblock \href {https://arxiv.org/abs/1707.06347} {Proximal policy optimization algorithms}.
\newblock \emph{arXiv preprint arXiv:1707.06347}.

\bibitem[{Shao et~al.(2024)Shao, Wang, Zhu, Xu, Song, Bi, Zhang, Zhang, Li, Wu et~al.}]{shao2024deepseekmath}
Zhihong Shao, Peiyi Wang, Qihao Zhu, Runxin Xu, Junxiao Song, Xiao Bi, Haowei Zhang, Mingchuan Zhang, YK~Li, Y~Wu, and 1 others. 2024.
\newblock \href {https://arxiv.org/abs/2502.03300} {Deepseekmath: Pushing the limits of mathematical reasoning in open language models}.
\newblock \emph{arXiv preprint arXiv:2402.03300}.

\bibitem[{Sheng et~al.(2025)Sheng, Zhang, Ye, Wu, Zhang, Zhang, Peng, Lin, and Wu}]{verl}
Guangming Sheng, Chi Zhang, Zilingfeng Ye, Xibin Wu, Wang Zhang, Ru~Zhang, Yanghua Peng, Haibin Lin, and Chuan Wu. 2025.
\newblock Hybridflow: A flexible and efficient rlhf framework.
\newblock In \emph{Proceedings of the Twentieth European Conference on Computer Systems}, pages 1279--1297.

\bibitem[{Team et~al.(2025{\natexlab{a}})Team, Kamath, Ferret, Pathak, Vieillard, Merhej, Perrin, Matejovicova, Ram{\'e}, Rivi{\`e}re et~al.}]{gemma3report}
Gemma Team, Aishwarya Kamath, Johan Ferret, Shreya Pathak, Nino Vieillard, Ramona Merhej, Sarah Perrin, Tatiana Matejovicova, Alexandre Ram{\'e}, Morgane Rivi{\`e}re, and 1 others. 2025{\natexlab{a}}.
\newblock \href {https://arxiv.org/abs/2503.19786} {Gemma 3 technical report}.
\newblock \emph{arXiv preprint arXiv:2503.19786}.

\bibitem[{Team et~al.(2025{\natexlab{b}})Team, Du, Gao, Xing, Jiang, Chen, Li, Xiao, Du, Liao et~al.}]{kimik15}
Kimi Team, Angang Du, Bofei Gao, Bowei Xing, Changjiu Jiang, Cheng Chen, Cheng Li, Chenjun Xiao, Chenzhuang Du, Chonghua Liao, and 1 others. 2025{\natexlab{b}}.
\newblock \href {https://arxiv.org/abs/2501.12599} {Kimi k1. 5: Scaling reinforcement learning with llms}.
\newblock \emph{arXiv preprint arXiv:2501.12599}.

\bibitem[{Team(2024)}]{qwq-32b-preview}
Qwen Team. 2024.
\newblock \href {https://qwenlm.github.io/blog/qwq-32b-preview/} {Qwq: Reflect deeply on the boundaries of the unknown}.

\bibitem[{Wen et~al.(2025)Wen, Cai, Xiao, He, An, Duan, Du, Liu, Tang, Lv et~al.}]{wen2025light}
Liang Wen, Yunke Cai, Fenrui Xiao, Xin He, Qi~An, Zhenyu Duan, Yimin Du, Junchen Liu, Lifu Tang, Xiaowei Lv, and 1 others. 2025.
\newblock \href {https://arxiv.org/abs/2503.10460} {Light-r1: Curriculum sft, dpo and rl for long cot from scratch and beyond}.
\newblock \emph{arXiv preprint arXiv:2503.10460}.

\bibitem[{Wu et~al.(2025)Wu, Wang, Du, Jegelka, and Wang}]{wu2025more}
Yuyang Wu, Yifei Wang, Tianqi Du, Stefanie Jegelka, and Yisen Wang. 2025.
\newblock \href {https://arxiv.org/abs/2502.07226} {When more is less: Understanding chain-of-thought length in llms}.
\newblock \emph{arXiv preprint arXiv:2502.07266}.

\bibitem[{Xiang et~al.(2025)Xiang, Snell, Gandhi, Albalak, Singh, Blagden, Phung, Rafailov, Lile, Mahan et~al.}]{xiang2025towards}
Violet Xiang, Charlie Snell, Kanishk Gandhi, Alon Albalak, Anikait Singh, Chase Blagden, Duy Phung, Rafael Rafailov, Nathan Lile, Dakota Mahan, and 1 others. 2025.
\newblock \href {https://arxiv.org/abs/2501.04682} {Towards system 2 reasoning in llms: Learning how to think with meta chain-of-though}.
\newblock \emph{arXiv preprint arXiv:2501.04682}.

\bibitem[{Xie et~al.(2025)Xie, Gao, Ren, Luo, Hong, Dai, Zhou, Qiu, Wu, and Luo}]{xie2025logic}
Tian Xie, Zitian Gao, Qingnan Ren, Haoming Luo, Yuqian Hong, Bryan Dai, Joey Zhou, Kai Qiu, Zhirong Wu, and Chong Luo. 2025.
\newblock \href {https://arxiv.org/abs/2502.14768} {Logic-rl: Unleashing llm reasoning with rule-based reinforcement learning}.
\newblock \emph{arXiv preprint arXiv:2502.14768}.

\bibitem[{Xu et~al.(2025)Xu, Wu, Wang, Li, Zheng, Chen, Hu, Kang, Ji, Zhang et~al.}]{xu2025redstar}
Haotian Xu, Xing Wu, Weinong Wang, Zhongzhi Li, Da~Zheng, Boyuan Chen, Yi~Hu, Shijia Kang, Jiaming Ji, Yingying Zhang, and 1 others. 2025.
\newblock \href {https://arxiv.org/abs/2501.11284} {Redstar: Does scaling long-cot data unlock better slow-reasoning systems?}
\newblock \emph{arXiv preprint arXiv:2501.11284}.

\bibitem[{Yang et~al.(2024)Yang, Yang, Zhang, Hui, Zheng, Yu, Li, Liu, Huang, Wei et~al.}]{qwen25report}
An~Yang, Baosong Yang, Beichen Zhang, Binyuan Hui, Bo~Zheng, Bowen Yu, Chengyuan Li, Dayiheng Liu, Fei Huang, Haoran Wei, and 1 others. 2024.
\newblock \href {https://arxiv.org/abs/2412.15115} {Qwen2. 5 technical report}.
\newblock \emph{arXiv preprint arXiv:2412.15115}.

\bibitem[{Ye et~al.(2025)Ye, Huang, Xiao, Chern, Xia, and Liu}]{ye2025limo}
Yixin Ye, Zhen Huang, Yang Xiao, Ethan Chern, Shijie Xia, and Pengfei Liu. 2025.
\newblock \href {https://arxiv.org/abs/2502.03387} {Limo: Less is more for reasoning}.
\newblock \emph{arXiv preprint arXiv:2502.03387}.

\bibitem[{Yeo et~al.(2025)Yeo, Tong, Niu, Neubig, and Yue}]{yeo2025demystifying}
Edward Yeo, Yuxuan Tong, Morry Niu, Graham Neubig, and Xiang Yue. 2025.
\newblock \href {https://arxiv.org/abs/2502.03373} {Demystifying long chain-of-thought reasoning in llms}.
\newblock \emph{arXiv preprint arXiv:2502.03373}.

\bibitem[{Yu et~al.(2024)Yu, Xu, Weston, and Kulikov}]{yu2024distilling}
Ping Yu, Jing Xu, Jason~E Weston, and Ilia Kulikov. 2024.
\newblock Distilling system 2 into system 1.
\newblock In \emph{The First Workshop on System-2 Reasoning at Scale, NeurIPS'24}.

\bibitem[{Zeng et~al.(2025)Zeng, Huang, Liu, Liu, He, Ma, and He}]{simplerl}
Weihao Zeng, Yuzhen Huang, Qian Liu, Wei Liu, Keqing He, Zejun Ma, and Junxian He. 2025.
\newblock \href {https://arxiv.org/abs/2503.18892} {Simplerl-zoo: Investigating and taming zero reinforcement learning for open base models in the wild}.
\newblock \emph{arXiv preprint arXiv:2503.18892}.

\bibitem[{Zheng et~al.(2024)Zheng, Zhang, Zhang, Ye, and Luo}]{llamafactory}
Yaowei Zheng, Richong Zhang, Junhao Zhang, Yanhan Ye, and Zheyan Luo. 2024.
\newblock \href {https://doi.org/10.18653/v1/2024.acl-demos.38} {{L}lama{F}actory: Unified efficient fine-tuning of 100+ language models}.
\newblock In \emph{Proceedings of ACL}.

\end{thebibliography}

\appendix

\section{Detailed Experimental Setups}
\label{appendix:detailed_setup}

\subsection{Models}
\begin{table}[ht]
    \centering
    \small
    \renewcommand{\arraystretch}{1.2}
    \begin{tabular}{ll}
        \toprule
        \textbf{Category} & \textbf{Models} \\
        \midrule
        \textit{Qwen Family} & Qwen2.5-0.5B-Instruct,\\& Qwen2.5-1.5B-Instruct, \\
        & Qwen2.5-3B-Instruct, \\&Qwen2.5-7B-Instruct, \\
        & Qwen2.5-14B-Instruct \\
        \textit{Llama Family} & Llama3.2-1B-Instruct,\\& Llama3.2-3B-Instruct, \\
        & Llama3.1-8B-Instruct \\
        \textit{Gemma Family} & Gemma3-1B-IT \\
        \bottomrule
    \end{tabular}
    \caption{Overview of models investigated in this work.}
    \label{tab:models_overview}
\end{table}

Table~\ref{tab:models_overview} summarizes all models evaluated in this study.

\subsection{Supervised Fine-Tuning}
\label{appendix_subsec:sft_training_setup}

Supervised Fine-Tuning (SFT) is conducted on a dataset $\mathcal{D} = \{(x^{(i)}, y^{(i)})\}_{i=1}^N$, where each prompt $x^{(i)}$ is paired with an output $y^{(i)}$, which may include a long CoT.
The objective is to maximize the conditional log-likelihood $\log p_\theta(y^{(i)} \mid x^{(i)})$, encouraging the model to reproduce high-quality responses with structured reasoning.

\begin{table}[ht]
    \centering
    \small
    \renewcommand{\arraystretch}{1.2}
    \resizebox{\columnwidth}{!}{
    \begin{tabular}{lcccccc}
    \toprule
    \multirow{2}{*}{\textbf{Hyper-parameter}} & \multicolumn{6}{c}{\textbf{Long CoT Data Amount}}                                        \\[0.5ex] \cline{2-7} \noalign{\vskip 0.7ex}
                                              & \textbf{8k} & \textbf{16k} & \textbf{32k} & \textbf{64k} & \textbf{128k} & \textbf{220k} \\ \midrule
    Number of Epochs                          & 4           & 4            & 4            & 3            & 3             & 2             \\
    Batch Size                                & 8           & 16           & 32           & 64           & 128           & 128           \\
    Learning Rate                             & \multicolumn{6}{c}{$5\times 10^{-5}$}                                                    \\
    Optimizer                                 & \multicolumn{6}{c}{Adamw}                                                                \\
    Learning Rate Scheduler                   & \multicolumn{6}{c}{cosine}                                                               \\
    Max Sequence Length                       & \multicolumn{6}{c}{16384}                                                                \\
    Warmup ratio                              & \multicolumn{6}{c}{0.05}                                                                 \\
    Training Precision                        & \multicolumn{6}{c}{bfloat16}                                                             \\ \bottomrule
    \end{tabular}
    }
    \caption{Hyperparameters used for full-parameter supervised fine-tuning.}
    \label{tab:sft_hyperparameter}
\end{table}

\paragraph{Training Setup.}
Our SFT training is conducted using LLaMA-Factory~\cite{llamafactory} on a server equipped with 8 H100 (80GB) GPUs. The SFT experiments consumed approximately 2,500 GPU hours in total. We adopt full-parameter fine-tuning for all SFT experiments. The detailed hyperparameters we used are presented in Table \ref{tab:sft_hyperparameter}, which are determined through a preliminary hyperparameter search.

\paragraph{Evaluation Setup.}
We conduct evaluation using the official Qwen2.5-Math repository\footnote{\url{https://github.com/QwenLM/Qwen2.5-Math}} and UltraEval~\cite{he2024ultraeval}. Notably, we leverage UltraEval's vLLM-based multi-GPU, data parallel deployment to accelerate inference~\cite{kwon2023efficient}.

\subsection{Reinforcement Learning}
\label{appendix_rl_training_setup}

DeepSeek-R1~\cite{deepseek-r1} conducts large-scale RL using long CoT supervised fine-tuning as a cold start, establishing a widely adopted training pipeline for reasoning-oriented models. In this work, we adopt the same setup to study how long CoT supervision during SFT influences model behavior in the subsequent RL stage.

For the RL algorithm, we use GRPO~\cite{shao2024deepseekmath}, a computationally efficient variant of PPO~\cite{schulman2017proximal} that eliminates the need for a separate value model by estimating advantages using group-normalized rewards.

In line with DeepSeek-R1 and similar works, we employ a rule-based binary reward function: the model receives a reward of 1 if the final answer is correct, and 0 otherwise. This simple yet effective setup allows us to isolate the effect of long CoT SFT on the optimization behavior during RL.

\begin{table}[ht]
    \centering
    \small
    \renewcommand{\arraystretch}{1.2}
    \begin{tabular}{lc}
        \toprule
        \textbf{Hyper-parameter} & \textbf{Value} \\
        \midrule
        Training Algorithm & GRPO \\
        Prompt Batch Size & 1024 \\
        Rollout Per Prompt & 8 \\
        Maximum Rollout Length & 8192 \\
        Mini-Batch Size & 256 \\
        Sampling Temperature & 1.0 \\
        KL Loss Coefficient & $1\times 10^{-4}$ \\
        Learning Rate & $5\times 10^{-7}$ \\
        \bottomrule
    \end{tabular}
    \caption{Hyperparameters used for RL training.}
    \label{tab:rl_hyperparameter}
\end{table}

\paragraph{Training Setup.}
Our RL training is conducted using verl~\cite{verl} framework, on a server equipped with eight H100 GPUs (each with 80GB of memory). The RL experiments consumed approximately 5,000 GPU hours in total. The detailed hyperparameters used in our experiments are shown in Table~\ref{tab:rl_hyperparameter}.

\subsection{Reflection Behavior Analysis Setup}
\label{appendix_reflection_setup}
We adopt a cross-validation approach, following the methodology proposed in~\cite{drgrpo}, combining both keyword detection and LLM-based identification to detect the self-reflective behaviors in model outputs. In particular, the keyword pool used in this work is limited to: recheck,  rethink, reassess, reevaluate, re-evaluate, reevaluation, re-examine, reexamine, reconsider,  reanalyze, double-check, check again, think again, verify again, and go over the steps.

\begin{figure*}[ht]
    \centering
    \includegraphics[width=1\linewidth]{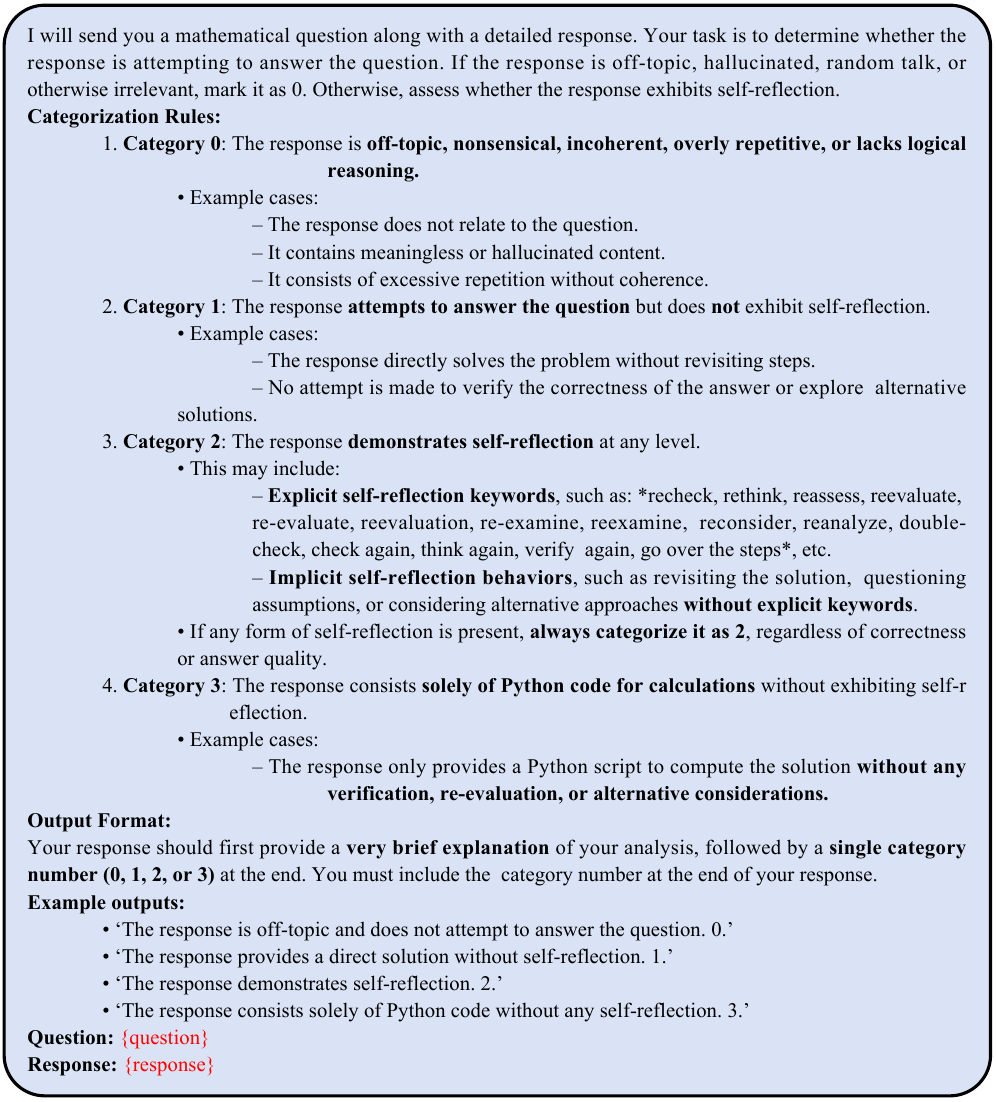}
    \caption{Prompt template used by GPT-4o-mini for reflection behavior identification.}
    \label{fig:gpt4o_prompt}
\end{figure*}

Figure \ref{fig:gpt4o_prompt} presents the prompt employed to determine whether a response contains self-reflection behaviors with an LLM-based approach.

\subsection{Arithmetic Benchmark}
\label{appendix_arithmetic_example}
% Figure \ref{fig:arithmetic_answer} shows an example from our synthetic arithmetic benchmark.
\paragraph{Qualitative Examples.}
To complement our quantitative analysis, we present three representative examples in Figures~\ref{fig:appendix_baseline}, \ref{fig:appendix_8k}, and \ref{fig:appendix_64k}. 
These responses are generated by Qwen2.5-3B-Instruct models under different training settings: the baseline model, the model finetuned on 8k-length CoT data, and the model finetuned on 64k-length CoT data, respectively. The baseline model solves the problem correctly with a clear and concise reasoning process. In contrast, the 8k-finetuned model produces a much longer and more verbose response. Despite proposing multiple alternative solutions, it repeatedly makes the same arithmetic mistakes, which propagate across steps and compound the final error. The response also exhibits disfluency, including repetitive phrases like “wait” and redundant equation restatements. The 64k-finetuned model shows significant improvement: it maintains more coherent structure, reflects on its steps more effectively, and proposes alternatives in a way that leads to the correct solution. These examples reinforce our hypothesis that longer responses introduce more room for error and noise—unless offset by sufficient training on long-form reasoning patterns.
\section{Detailed Evaluation Results}
\label{appendix:detailed_evaluation_results}

\subsection{Long CoT Degradation}
\label{appendix:subsec:long_cot_degradation}

\begin{figure*}[ht]
    \centering
    \includegraphics[width=1\linewidth]{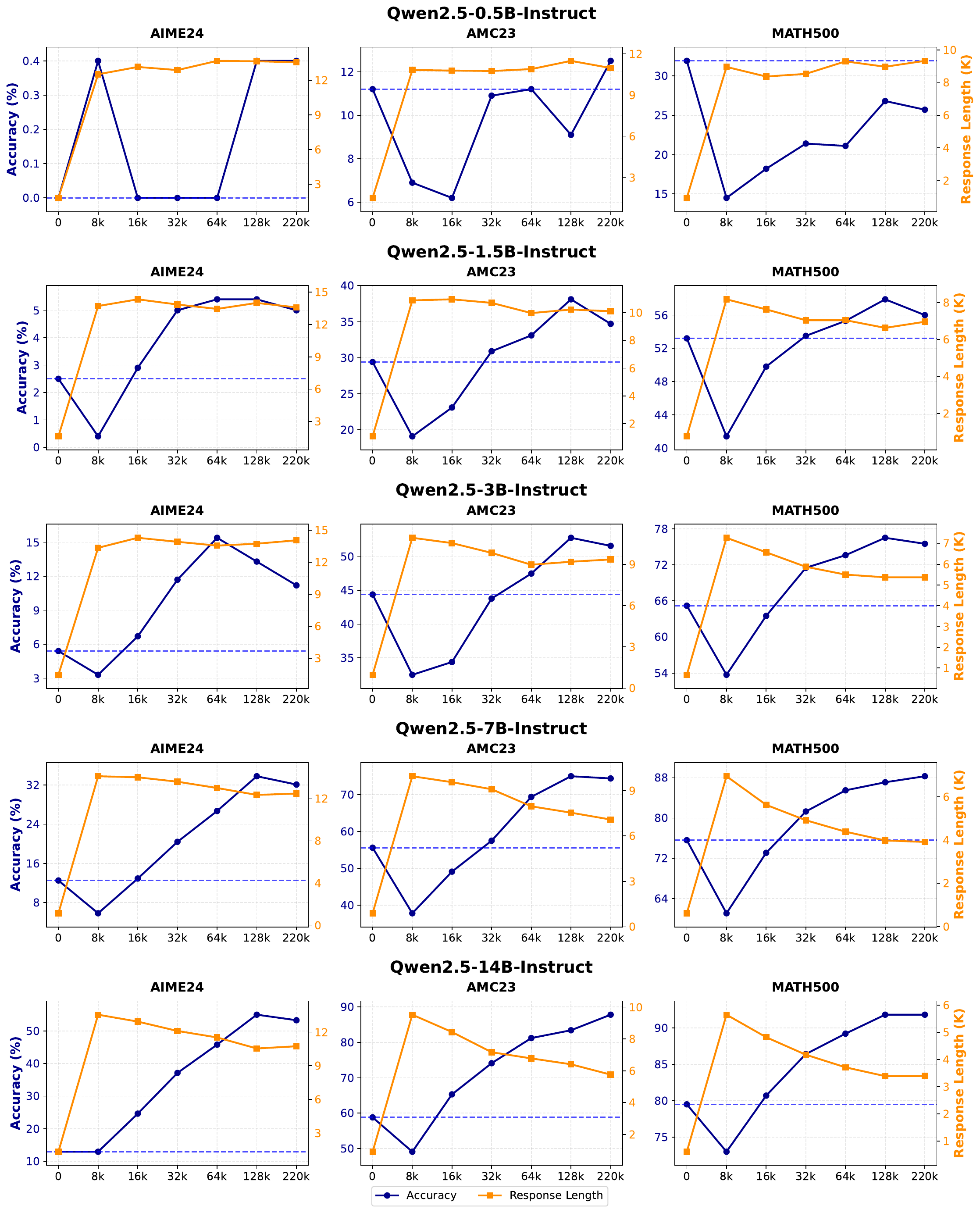}
    \caption{Comprehensive evaluation of Qwen models trained with varying amounts of long CoT data.}
    \label{fig:qwen_model_all_benchmark_acc_length}
\end{figure*}

\begin{figure*}[ht]
    \centering
    \includegraphics[width=1\linewidth]{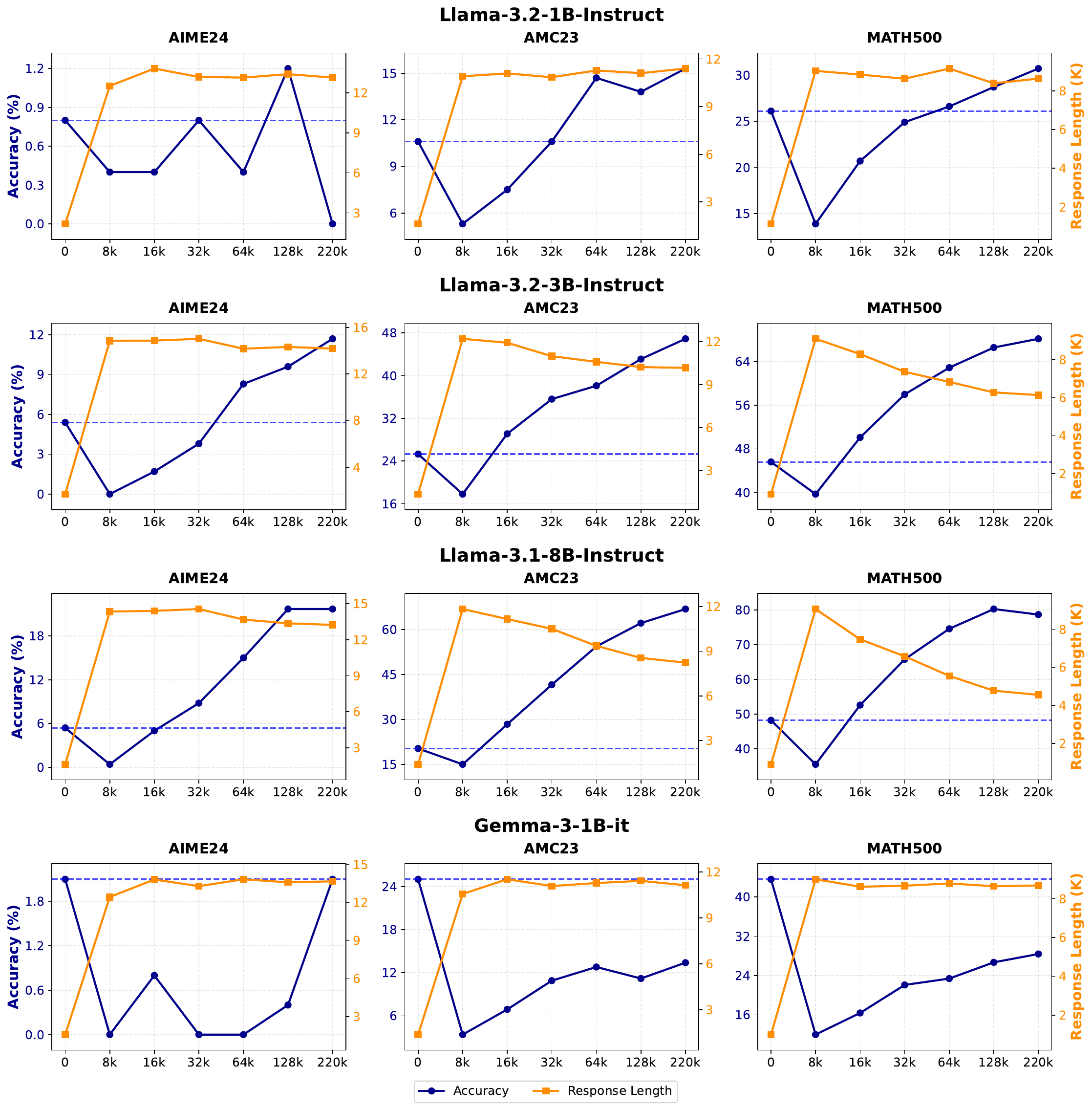}
    \caption{Comprehensive evaluation of LLaMA and Gemma models trained with varying amounts of long CoT data.}
    \label{fig:llama_gemma_model_all_benchmark_acc_length}
\end{figure*}

Figure \ref{fig:qwen_model_all_benchmark_acc_length} and \ref{fig:llama_gemma_model_all_benchmark_acc_length} show more detailed evaluation results for the experiments in Section \ref{sec:Long_CoT_Degradation}.

\begin{figure}[ht]
    \centering
    \includegraphics[width=1\linewidth]{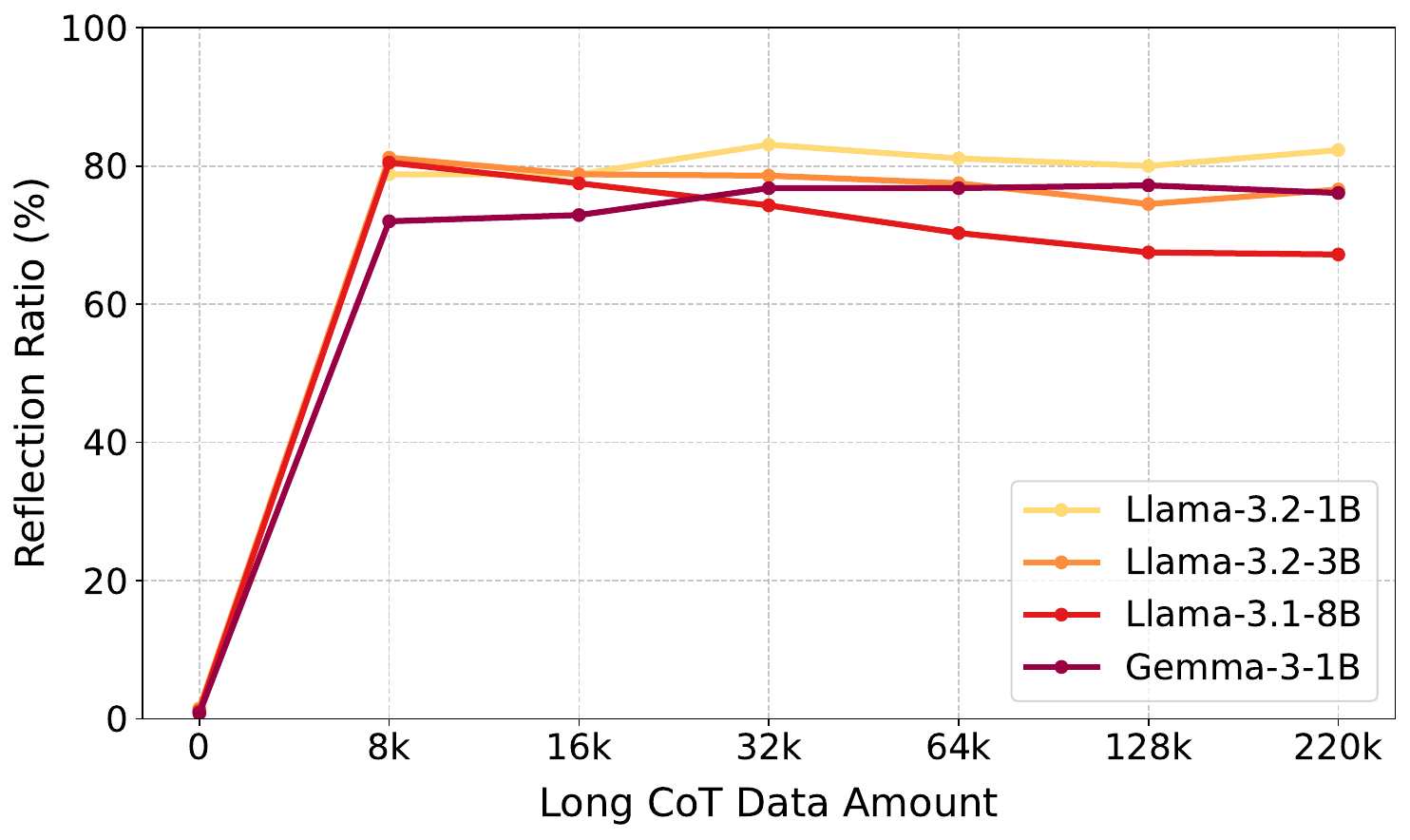}
    \caption{Reflection ratios of LLaMA and Gemma models of different sizes trained on varying amounts of long CoT data. The reflection ratio refers to the proportion of model responses (out of 4,000 on the MATH500 benchmark) that exhibit reflective behavior, as identified through cross-validation.}
    \label{fig:llama_gemma_reflection_ratio}
\end{figure}

\subsection{Reflection Behavior Analysis}
\label{appendix:subsec:reflection_analysis}

\begin{figure*}[ht]
    \centering
    \includegraphics[width=1\linewidth]{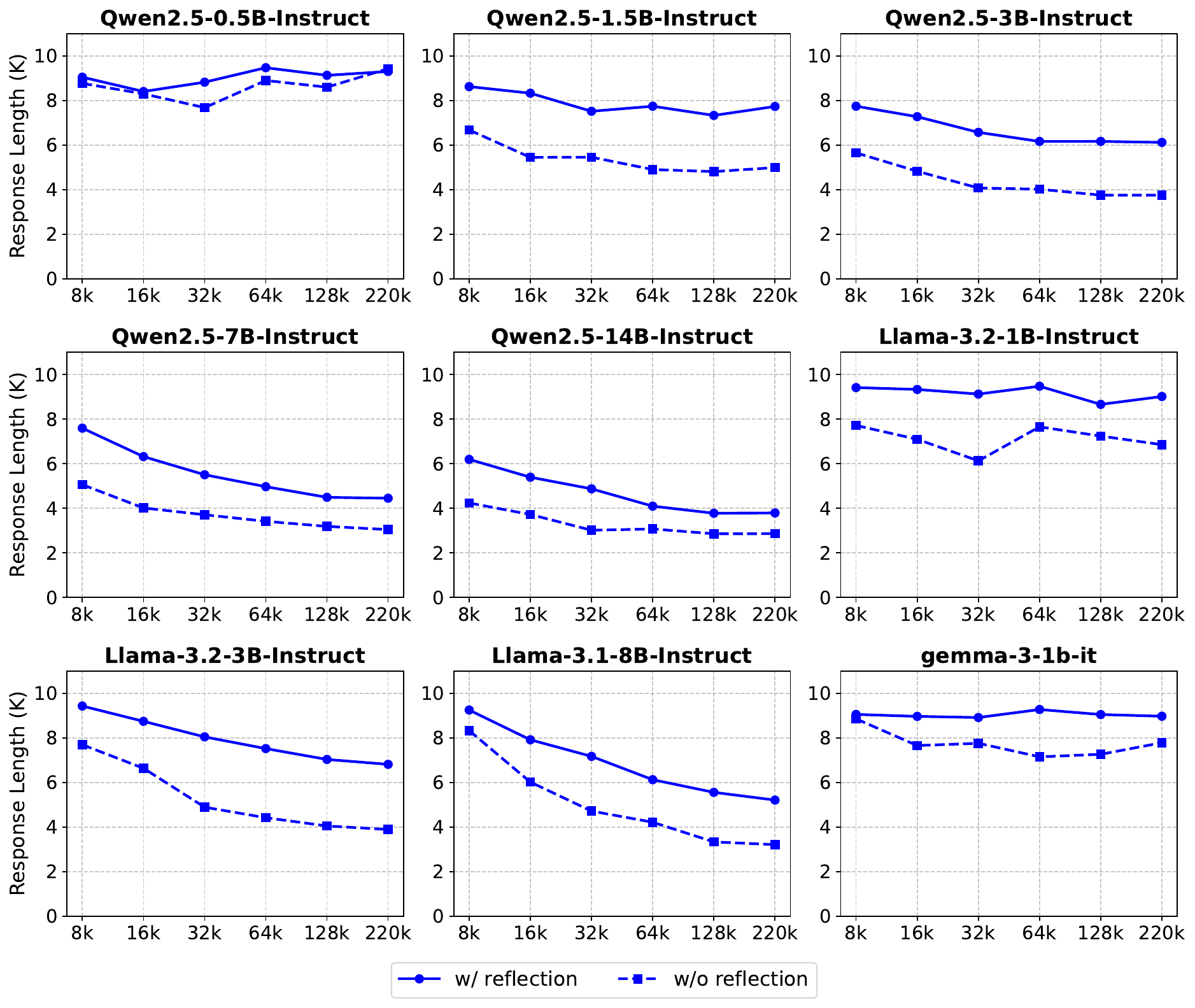}
    \caption{Average response lengths of multiple models trained with varying amounts of long CoT data. Solid lines represent responses exhibiting reflection behavior; dashed lines denote responses without reflection.}
    \label{fig:all_model_reflection_length_result}
\end{figure*}

Figure \ref{fig:llama_gemma_reflection_ratio} and \ref{fig:all_model_reflection_length_result} are the detailed evaluation results for the experiments in Section \ref{subsec:reflection_behavior_analysis}.

\definecolor{contextgray}{gray}{0.96}
\definecolor{instructionblue}{rgb}{0.9,0.95,1.0}
\definecolor{responsegreen}{rgb}{0.94,1.0,0.94}

\newtcolorbox{examplebox}[1]{colback=white, colframe=black!50, title=\textbf{Example #1}, fonttitle=\bfseries, arc=2pt, boxrule=0.4pt, top=1mm, bottom=1mm}

\newtcolorbox{innerbox}[2]{colback=#1, colframe=black!40, boxrule=0.3pt, arc=2pt, title=#2, left=1mm, right=1mm, top=0.5mm, bottom=0.5mm, fonttitle=\bfseries}

\begin{figure*}[t]  % Or [b], [h], etc.
\centering
\begin{examplebox}{1 Qwen2.5-3B-Instruct}

\begin{innerbox}{contextgray}{Question}
\small
Compute the answer to this arithmetic expression: (14*21*38+32+23+31+3)-11+59+71+60
\end{innerbox}

\begin{innerbox}{responsegreen}{Response}
\small
To solve the arithmetic expression \((14 \times 21 \times 38 + 32 + 23 + 31 + 3) - 11 + 59 + 71 + 60\), we need to follow the order of operations, which means we should perform the multiplication before the addition and subtraction.

First, calculate the product \(14 \times 21 \times 38\):

\[
14 \times 21 = 294
\]
\[
294 \times 38 = 11172
\]

Now substitute this back into the expression:

\[
11172 + 32 + 23 + 31 + 3 - 11 + 59 + 71 + 60
\]

Next, add the numbers inside the parentheses:

\[
11172 + 32 + 23 + 31 + 3 = 11261
\]

Now, substitute this back into the expression:

\[
11261 - 11 + 59 + 71 + 60
\]

Perform the subtraction and addition from left to right:

\[
11261 - 11 = 11250
\]
\[
11250 + 59 = 11309
\]
\[
11309 + 71 = 11380
\]
\[
11380 + 60 = 11440
\]

Thus, the final answer is:

\[
\boxed{11440}
\]
\end{innerbox}

\end{examplebox}

\caption{Example from our synthetic arithmetic benchmark. The response is generated by the Qwen2.5-3B-Instruct model. The model follows a step-by-step approach, applying the order of operations to solve the arithmetic expression. }
\label{fig:appendix_baseline}
\end{figure*}

\begin{figure*}[t]  % Or [b], [h], etc.
\centering
\begin{examplebox}{2 Qwen2.5-3B-Instruct 8k}

\begin{innerbox}{contextgray}{Question}
\small
Compute the answer to this arithmetic expression: (14*21*38+32+23+31+3)-11+59+71+60
\end{innerbox}

\begin{innerbox}{responsegreen}{Response}
\small
Okay, let's see. I need to compute this arithmetic expression: (14*21*38 + 32 + 23 + 31 + 3) - 11 + 59 + 71 + 60. Hmm, that looks a bit complicated, but let's break it down step by step. 

First, I should handle the operations inside the parentheses before dealing with the numbers outside. The expression starts with 14 multiplied by 21 multiplied by 38. Let me calculate that first. Multiplying those numbers might take some time, so maybe I can do them step by step. Let's see, 14 times 21. Hmm, 14*20 is 280, and 14*1 is 14, so 280+14=294. Then, multiplying that result by 38. Okay, 294*38. Let me break that down too. Maybe 294*30 is 8820, and 294*8 is 2352. Adding those together, 8820+2352. Let me add them step by step: 8820 + 2000 is 10820, then +352 more is 11172. So, 14*21*38 equals 11172.

Next, the problem adds 32 to that result. So, 11172 + 32. That's straightforward, 11172 + 30 is 11202, plus 2 more is 11204. Then, it adds 23: 11204 +23. Let's see, 11204 +20 is 11224, plus 3 is 11227. Next, adding 31: 11227 +31. 11227 +30 is 11257, plus 1 more is 11258. Then, adding 3: 11258 +3. That's just 11261. So, inside the parentheses simplifies to 11261.

Now, the expression continues with minus 11: 11261 -11. That's easy, 11261 -10 is 11251, minus 1 more is 11250. Then, adding 59: 11250 +59. Let me check that. \textcolor{red}{11250 +50 is 11290, plus 9 is 11299.} Next, adding 71: 11299 +70 is 11369, plus 1 is 11370. Then, adding 60: 11370 +60. That's 11370 +60. Let me verify. 11370 +60: \textcolor{red}{11370 +50 is 11320, plus 10 is 11330, plus 10 is 11340.} Wait, no, hold on. Wait, 11370 +60: 11370 +60. Let me do this differently. 11370 + 60 is the same as 11370 + 6*10. So, 11370 +60 = 11370 + 6*10. 11370 +60 is 11370 + 60. Alternatively, 11370 + 60 = 11370 + 60. Let me add 11370 + 60. \textcolor{red}{11370 + 60: 11370 + 50 is 11320, then +10 is 11330.} Wait, that doesn't seem right. Wait, 11370 +60: 11370 + 60. Let's think of 11370 as 1137*10. Then 1137*10 +6*10 = 11370 +60. Alternatively, 11370 +60: 11370 + 60. Let's do 11370 + 60. \textcolor{red}{11370 + 50 = 11320, then +10 = 11330.} Wait, but 11370 +60 is actually 11370 + 6*10. So, 11370 +60 = 11370 + 60. Let me verify with another approach. 11370 +60: 11370 + 60. If I break down 60 into 50 +10, \textcolor{red}{then 11370 +50 = 11320, then +10 = 11330.} Alternatively, 11370 +60: 11370 + 60. Let me do 11370 + 60. Let's count from 11370: adding 10 is 11380, adding another 10 is 11390. Wait, no. Wait, 11370 +60: 11370 + 60. Let me use column addition. 

......

% 11370 is 1137*10. Adding 60 (which is 6*10) gives 1137*10 +6*10 = (1137 +6)*10 = 1143*10 = 11430.ß Wait, that can't be right. Wait, 11370 +60. Let me check with another method. 11370 +60: If I move the decimal point two places to the left, 11370 becomes 113.7, then moving it back, it's 11430. Wait, that seems conflicting. Wait, no. Wait, 11370 +60. Let me use the standard addition algorithm. 

% 11370
% +   60
% --------
% 11430

% Wait, that's not possible. Wait, 11370 +60. Let's add 11370 +60. If I write 11370 as 1137*10, then adding 60 is adding 6*10, so total is (1137 +6)*10 = 1143*10 = 11430. But when I thought of it as 11370 +60, moving the decimal, it's 11430. Wait, but 11370 +60. Let me verify with actual numbers. Let's take 11370 +60. If I add 11370 +60. Let's break it down as 11370 +60. 11370 + 60. If I add 11370 +60. Let me do 11370 +60. 11370 + 60. Let's think of 11370 as 1137*10. So, 1137*10 +6*10 = 11370 +60. So, 11370 +60 = 11430. But when I add 11370 +60, using column addition:

% 11370
% +   60
% -------
% 11430

Wait, that's correct. So, 11370 +60 = 11430. So, in that case, 11370 +60 = 11430. So, 11430 is the sum of 11370 and 60. Okay, so that's correct. Therefore, 11261 -11 +59 +71 +60 equals 11261 -11 + (59+71+60). Let me compute 59+71 first. 59+71: 50+70 is 120, 9+1 is 10, so 120+10=130. Then, 130+60 is 190. So, 59+71+60=190. Then, 11261 -11 is 11250, plus 190 is 11250 +190. Let me add those. \textcolor{red}{11250 +190: 11250 +100 is 11350, then +90 is 11350 +90=11340.} So, the total is 11340. 

Wait, let me check that again. 11261 -11 is 11261 -10=11251, minus 1=11250. Then, 11250 +59=11250+59. Let's do \textcolor{red}{11250 +50=11290, then +9=11299.} Then, 11299 +71=11299+70=11369, +1=11370. Then, 11370 +60=11430. So, total is 11430. 

Alternatively, maybe there's a different way to compute the total without breaking it down so many times. Let me see. The original expression is (14*21*38 +32+23+31+3) -11+59+71+60. So, the first part is 14*21*38 + (32+23+31+3). 

...... 

So, that's 11261. Then, 11261 -11=11250, +59=11250+59=\textcolor{red}{11250+50+9=11290+9=11299.} Then, +71=11299+70=11369+1=11370. Then, +60=11370+60=11430. So, total is 11430. 

......

Wait, maybe the error is in the initial step of adding the numbers inside the parentheses. Let me check that again. The first part is (14*21*38 +32 +23 +31 +3). Let's compute each term:

14*21*38: 14*21=294, 294*38=11172. Correct.

......

So, inside the parentheses is 11261. Then, subtract 11: 11261 -11=11250. Correct.

Then, add 59: \textcolor{red}{11250 +59=11250 +50 +9=11290 +9=11299.} Correct.

Add 71: 11299 +71=11299 +70 +1=11369 +1=11370. Correct.

Add 60: 11370 +60=11370 +60=11430. Correct.

So, the total is 11430. Therefore, the first method is correct. Then, why does the second method give a different result? Because in the second method, the user added the constants first.
......

Therefore, the total is 11430. Therefore, the second method is incorrect because it misapplies the order of operations. The user added the constants first, which is not the case in the original expression. The original expression requires the constants to be added after the product is computed and before the subtraction and additions. Therefore, the correct total is 11430.

Therefore, the final answer is \boxed{11430}.

\end{innerbox}

\end{examplebox}

\caption{Example from our synthetic arithmetic benchmark. The response is generated by Qwen2.5-3B-Instruct finetuned on 8k long CoT data. Despite proposing multiple alternative solutions, the model makes similar calculation errors, which propagate through subsequent steps and result in compounding mistakes. Incorrect steps are highlighted in \textcolor{red}{red}. The response is truncated as it exceeds 5,000 tokens.}
\label{fig:appendix_8k}
\end{figure*}

\begin{figure*}[t]  % Or [b], [h], etc.
\centering
\begin{examplebox}{3 Qwen2.5-3B-Instruct 64k}

\begin{innerbox}{contextgray}{Question}
\small
Compute the answer to this arithmetic expression: (14*21*38+32+23+31+3)-11+59+71+60
\end{innerbox}

\begin{innerbox}{responsegreen}{Response}
\small
Okay, so I need to compute this arithmetic expression: (14*21*38 + 32 + 23 + 31 + 3) - 11 + 59 + 71 + 60. Hmm, let me break this down step by step. First, I should handle the operations inside the parentheses, which is the product of 14, 21, and 38, then add the other numbers. Let me start with that.

Calculating 14 multiplied by 21. Let me do 14*20 first, which is 280, and then add 14*1, which is 14. So 280 + 14 = 294. Okay, so 14*21 is 294. Now, I need to multiply that result by 38. Hmm, 294*38. Let me break that down. Maybe 294*40 would be easier, but subtract 294*2. Wait, 294*40 is 11,760 and 294*2 is 588. So 11,760 - 588 = 11,172. So 14*21*38 is 11,172. \textcolor{blue}{Let me check that again to be sure. 294*38: 294*30 is 8,820 and 294*8 is 2,352. Adding those together: 8,820 + 2,352. Let's see, 8,000 + 2,000 is 10,000, 820 + 352 is 1,172. So 10,000 + 1,172 is 11,172. Yep, that matches. So that part is correct.}

Now, add the numbers 32, 23, 31, and 3. Let's do that step by step. 32 + 23 is 55. Then 55 + 31 is 86. Then 86 + 3 is 89. So the sum inside the parentheses is 11,172 + 89. Let me compute that. 11,172 + 80 is 11,252, then +9 is 11,261. So the parentheses part equals 11,261.

Now the entire expression is 11,261 - 11 + 59 + 71 + 60. Let me handle the subtraction and addition from left to right. So first, 11,261 - 11. That's straightforward: 11,261 - 10 is 11,251, then subtract 1 more is 11,250. So now we have 11,250 + 59 + 71 + 60.

Let me add those numbers. 59 + 71. 50 + 70 is 120, 9 + 1 is 10, so total 130. Then 130 + 60 is 190. So adding 190 to 11,250 gives 11,250 + 190 = 11,440. \textcolor{blue}{Let me verify each step again to make sure I didn't make a mistake.}

\textcolor{blue}{First, 14*21*38: 14*21 is 294, 294*38 is 11,172. Then adding 32+23+31+3: 32+23=55, 55+31=86, 86+3=89. So 11,172 + 89 = 11,261. Then 11,261 - 11 = 11,250. Then adding 59+71+60: 59+71=130, 130+60=190. 11,250 + 190 = 11,440. Seems correct.}

Wait, let me check the addition again. 59 + 71: 59 + 70 is 129, plus 1 is 130. Then 130 + 60 is 190. Correct. Then 11,250 + 190: 11,250 + 100 is 11,350, plus 90 is 11,440. Yes. So the final answer should be 11,440.

\textcolor{blue}{But just to be thorough, let me compute the entire expression step by step again, maybe using a different order or verify the multiplication another way.}

\textcolor{blue}{Alternatively, maybe I can compute 14*21*38 again. Let's see, 14*21 is 294, then 294*38. Let's compute 294*38 as (300 - 6)*38. So 300*38 = 11,400, 6*38=228. So 11,400 - 228 = 11,172. Same result. Then adding 32+23+31+3: 32+23=55, 55+31=86, 86+3=89. So 11,172 + 89 = 11,261. Then subtract 11: 11,261 - 11 = 11,250. Then add 59,71,60. Let's do 59 + 71 first: 59 + 71. 59 + 70 is 129, plus 1 is 130. Then 130 + 60 is 190. So 11,250 + 190 is 11,440. All steps check out.}

Alternatively, maybe compute the entire expression as written:

(14*21*38 + 32 + 23 + 31 + 3) - 11 + 59 + 71 + 60

Let me compute all the additions first before subtracting:

First, compute the sum inside the parentheses: 14*21*38 + 32 + 23 + 31 + 3.

But we already did that. Then subtract 11, then add 59, 71, 60.

\textcolor{blue}{Alternatively, maybe compute each term step by step:}

Start with 14*21*38. Let's compute that again. 14*21 is 294. Then 294*38. Let me compute 294*38 as 294*(30 + 8) = 294*30 + 294*8. 294*30 is 8,820. 294*8 is 2,352. Adding those gives 8,820 + 2,352 = 11,172. Correct.

Then 32 + 23 + 31 + 3. Let's add them in pairs: 32 + 31 = 63, 23 + 3 = 26. Then 63 + 26 = 89. Correct. So the parentheses total 11,172 + 89 = 11,261.

Now subtract 11: 11,261 - 11 = 11,250. Then add 59, 71, 60. Let's add 59 + 71 first. 59 + 70 is 129, plus 1 is 130. Then 130 + 60 = 190. Then 11,250 + 190 = 11,440. All steps confirm the same result.

\textcolor{blue}{Alternatively, maybe compute the entire expression as written:}

14*21*38 + 32 + 23 + 31 + 3 - 11 + 59 + 71 + 60

But that's the same as before. \textcolor{blue}{Alternatively, compute each term in order:}

14*21*38 = 11,172

......

+60 = 11,440

Yes, same result. So regardless of the order, the answer is 11,440.

\textcolor{blue}{I think that's solid. Let me just check if I made any arithmetic errors in the final addition.} 11,250 + 190. 11,250 + 100 = 11,350. Then +90 is 11,440. Correct.

Alternatively, 190 is 100 + 90. 11,250 + 100 is 11,350. 11,350 + 90 is 11,440. Yep.

So I think that's the correct answer. 11,440. So the final answer is \boxed{11440}.

**Final Answer**
\boxed{11440}

\end{innerbox}

\end{examplebox}

\caption{Example from our synthetic arithmetic benchmark. Response is generated by Qwen2.5-3B-Instruct finetuned on 64k long CoT data. The model frequently engages in verification steps and occasionally proposes alternative solutions. Verification steps and alternative solution paths are highlighted in \textcolor{blue}{blue}.}
\label{fig:appendix_64k}
\end{figure*}

\end{document}